\documentclass[11pt]{article}
\usepackage[T1]{fontenc}
\usepackage[utf8]{inputenc}
\usepackage[letterpaper,margin=1in]{geometry}
\usepackage{graphicx}
\usepackage{tikz}
\usetikzlibrary{arrows.meta, positioning, shapes.geometric, calc, decorations.pathreplacing}
\usepackage[hyphens]{url}
\usepackage{booktabs}
\usepackage{amsmath}
\usepackage{array}
\usepackage{float}
\usepackage{placeins}
\usepackage{fvextra}
\usepackage[round]{natbib}
\usepackage[colorlinks=true,allcolors=blue]{hyperref}
\DefineVerbatimEnvironment{PromptBlock}{Verbatim}{breaklines=true,breakanywhere=false,breaksymbolleft={},breaksymbolright={},fontsize=\footnotesize}
\hypersetup{
  pdfauthor={Yihang Chen, Pin Qian, Su Wang, Chong Peng, Huan Xu, Shuaiting Li, Yiqi Sun},
  pdftitle={Explicit State Elicitation Is Not Enough: A Controlled Audit of Memory-Policy Classification},
  pdfsubject={Memory-policy classification audit},
  pdfkeywords={memory agents, policy classification, controlled audit, intermediate representations}
}

\title{Explicit State Elicitation Is Not Enough: A Controlled Audit of Memory-Policy Classification}

\author{%
\begin{tabular}{@{}c@{\quad}c@{\quad}c@{}}
Yihang Chen & Pin Qian & Su Wang \\
\footnotesize\href{mailto:ychen3726@gatech.edu}{\texttt{ychen3726@gatech.edu}} &
\footnotesize\href{mailto:pqian@alumni.cmu.edu}{\texttt{pqian@alumni.cmu.edu}} &
\footnotesize\href{mailto:suwang@alumni.cmu.edu}{\texttt{suwang@alumni.cmu.edu}} \\[0.5em]
Chong Peng & Huan Xu & Shuaiting Li \\
\footnotesize\href{mailto:chongp@alumni.cmu.edu}{\texttt{chongp@alumni.cmu.edu}} &
\footnotesize\href{mailto:huan.xu71@gmail.com}{\texttt{huan.xu71@gmail.com}} &
\footnotesize\href{mailto:list@zju.edu.cn}{\texttt{list@zju.edu.cn}} \\[0.5em]
\multicolumn{3}{c}{Yiqi Sun} \\
\multicolumn{3}{c}{\footnotesize\href{mailto:ysun697@gatech.edu}{\texttt{ysun697@gatech.edu}}}
\end{tabular}}
\date{}

\begin{document}
\maketitle

\begin{abstract}
Personalized agents must decide whether retrieved user memory should be used, ignored, updated, or queried before it affects a current task. We use this setting to develop an empirical audit protocol for structured intermediate outputs: first audit dataset shortcuts, then isolate bundled prompt changes, check whether intermediate labels are answer-associated, test decomposed semantic evidence, and audit provider-level execution failures. A 480-example synthetic development set initially suggested large gains from a state-structured prompt bundle, but TF-IDF diagnostics showed lexical separability and no positive standalone \textsc{Ignore} cases. We therefore construct a frozen 160-example controlled counterfactual set with 40 matched four-way families and rule-derived reference policies. On this set, exposing the four state definitions improves accuracy, but an isolated explicit state-output field does not significantly improve policy accuracy for Llama-3.3-70B and gives only a marginal, non-significant gain for GPT-OSS-120B. Supplying benchmark-associated state labels shifts policy predictions, but because those labels deterministically map to policies, this is a label-conditioning diagnostic rather than evidence of a faithful internal mechanism. Family-level and seed-stability analyses further show that example-level accuracy overstates counterfactual consistency: complete four-way family success is rare. An exploratory follow-up that elicits decomposed semantic evidence also fails to improve routing for the cleanly evaluated endpoint; the corresponding GPT-OSS condition was unavailable because of provider-side request validation. We evaluate policy classification only, not downstream responses, tool actions, or memory-store mutation.
\end{abstract}

\section{Introduction}
Long-term-memory agents retrieve user histories before answering current requests. Retrieval alone does not determine how the memory should be used: it may be current, irrelevant, superseded, or insufficient for the task. We study this intermediate decision as memory-policy classification over \{\textsc{Use}, \textsc{Ignore}, \textsc{Update}, \textsc{Ask}\}. The experiments stop at policy prediction; they do not evaluate final responses, tool actions, or memory-store mutation.

Our starting point is MemoryPolicy-Bench, a 480-example synthetic development set. It suggested that a state-structured prompt bundle could improve exact-match policy accuracy over implicit-state/no-state-output prompts on Llama-3.3-70B and Qwen3-32B. A later audit changed the interpretation: the development set is highly lexically separable under supervised TF-IDF diagnostics, and it positively instantiates \textsc{Use}, \textsc{Update}, and \textsc{Ask} but has no standalone positive \textsc{Ignore} reference class. We therefore treat it as development evidence and motivation, not as a decisive out-of-distribution test.

To test the prompt claim more directly, we freeze a 160-example controlled counterfactual set. It contains 40 scenario families, each with matched \textsc{Use}, \textsc{Ignore}, \textsc{Update}, and \textsc{Ask} variants. Reference policies are rule-derived from structured semantic relations before rendering. Each example has four history records, and the current task text is byte-identical across the four variants in a family. The set passes deterministic structural and semantic validation and was verified by two annotators, working separately and blind to the rule-derived reference policies, who agreed with each other and with the reference policy on all 160 examples; because the labels are deterministically rule-derived, this agreement partly reflects the construction. Under grouped diagnostics, the frozen controlled set remains harder for the evaluated lexical baselines, although this is not directly comparable to the original random-fold development-set diagnostic because the split, label space, and class priors differ. It is synthetic and rule-labeled.

The controlled results revise the paper's claim and motivate a broader audit framing. In a matched prompt ablation, asking the model to output an explicit state field does not significantly improve policy accuracy for Llama and is only marginal for GPT-OSS. A separate label-conditioning diagnostic shows that policy routing is sensitive to externally supplied benchmark-associated state labels. This diagnostic is intentionally reported narrowly: because states in the frozen controlled set map deterministically to policies, a supplied state is information-equivalent to an answer label for an oracle that knows the deterministic state-to-policy mapping; note, however, that the model does not route on it perfectly --- forced-correct-state accuracy is 52.7\% (Llama) and 65.0\% (GPT-OSS), since the forced-state prompt supplies the state but not the state-to-policy mapping. It does not establish that naturally emitted states are faithful internal mechanisms. We therefore analyze not only average accuracy but also family-level counterfactual consistency, stochastic seed stability, and rule-based error modes.

Our contributions are:
\begin{itemize}
    \item We propose a reusable audit protocol (instantiated here on memory-policy classification) that separates dataset shortcuts, bundled prompt changes, answer-associated label conditioning, semantic-evidence elicitation, and provider-level execution failures.
    \item We instantiate the protocol on memory-policy classification with matched four-way counterfactual families, family-cluster analysis, stochastic-stability diagnostics, and rule-based error-mode summaries.
    \item We show that apparent development-set state-structured gains do not survive an isolated state-output ablation, while supplied deterministic state labels should be interpreted as answer-associated conditioning rather than mechanism evidence.
\end{itemize}

\section{Related Work}
Long-term memory systems for LLM agents study how to store, organize, retrieve, and revise information across interactions. MemGPT treats memory as an operating-system-like hierarchy for moving information in and out of limited context windows \citep{packer2023memgpt}; A-MEM develops dynamic indexing, linking, and memory evolution \citep{xu2025amem}; LongMemEval evaluates extraction, multi-session reasoning, temporal reasoning, knowledge updates, and abstention in long-term chat memory \citep{wu2024longmemeval}; and LoCoMo evaluates very long-term conversational memory over multi-session dialogues \citep{maharana2024locomo}. PERSONAMEM studies dynamic user profiles and personalized responses at scale \citep{jiang2025personamem}, while Mem2ActBench evaluates long-term memory use in task-oriented autonomous-agent actions \citep{shen2026mem2actbench}. Recent multi-agent work also cautions that expanded recall can change downstream social behavior, so memory should be treated as an active determinant of agent behavior rather than a passive context extension \citep{liu2026memory}. These systems establish memory as an architectural requirement, whereas MemoryPolicy-Bench asks a complementary policy question: once a memory is retrieved, should it be used, queried, ignored, or updated?

Personalized language modeling and RAG benchmarks study how retrieved profile information supports outputs. LaMP evaluates personalized classification and generation with retrieval over user profiles \citep{salemi2024lamp}, PersonaLLM tests consistency under assigned personality traits \citep{jiang2024personallm}, and PrefEval evaluates whether models infer and follow user preferences in long-context conversations \citep{zhao2025prefeval}. PrefIx similarly evaluates preference-aware human-agent interaction quality alongside task success, rather than treating task accuracy as the only endpoint \citep{li2026prefixunderstandadaptuser}. Retrieval-Augmented Generation combines parametric models with retrieved evidence \citep{lewis2020retrieval}, but retrieved context can conflict with parametric knowledge, with other retrieved documents, or with itself \citep{xu2024knowledgeconflicts}; Self-RAG addresses part of this problem through adaptive retrieval and self-critique \citep{asai2024selfrag}. Other retrieval-stage work constructs semantically weighted path evidence for multi-hop knowledge-graph QA while preserving path-level traces for diagnostics \citep{Fu_2026}. Recent retrieval audits further show that evaluation can be confounded when retrieval units exceed an encoder's effective input window, and that query-adaptive fusion need not reliably outperform fixed fusion under grouped evaluation \citep{wu2026evidenceunitfairnesslimitsqueryadaptive}. Related benchmark work in regulatory divergence detection distinguishes agreement, divergence, and directional silence over paired source texts \citep{wu2026regdivergence}. That work studies regulatory-document comparison rather than user memory, but it illustrates why conflict and absence can require distinct labels. Related concerns arise in adaptive systems under distribution shift: DriftGuard uses multiple safety-relevant monitors to trigger selective adaptation for evolving toxicity moderation, rather than relying on global distribution drift alone \citep{xin2026driftguard}. These works are complementary to our setting: they study retrieval-stage evaluation, paired document comparison, or when a deployed model should adapt to changing input and risk distributions, whereas we audit how an agent should route retrieved user memory after retrieval.

Adjacent controlled evaluations outside memory agents make a similar methodological point: conclusions can depend on cohort and target definitions, on separating an intervention from confounded weighting changes, and on which metric is reported \citep{han2026earlyearlyenoughdesigndependent,wu2026classweightingversusconditioning,han2026interpretablevslearnedencoders}. We use these as methodological analogies for controlled evaluation design, not as evidence about memory-policy routing.

Clarification and calibration work motivate asking as a first-class action. CLAQUA supports deciding whether clarification is needed and how to ask it in knowledge-based QA \citep{xu2019asking}, CLAM shows that generative models can selectively ask clarifying questions for ambiguous inputs \citep{kuhn2023clam}, and self-knowledge work studies whether models know when they do not know \citep{kadavath2022know}. Knowledge-editing methods such as MEMIT directly alter model parameters to update stored facts \citep{meng2022memit}; MemoryPolicy-Bench instead leaves the model unchanged and tests whether a prompted agent can classify the policy for a retrieved user memory at decision time.

Prompting work shows that explicit intermediate text can change model behavior. Chain-of-thought prompting elicits free-form reasoning traces \citep{wei2022chain}, while Least-to-Most and Self-Ask decompose problems into intermediate subproblems or questions \citep{zhou2023least,press2023measuring}. More broadly, semantically equivalent prompt formulations can themselves induce systematic performance differences: Jin et al. characterize this phenomenon as prompt privilege and study accessibility-oriented prompt normalization to reduce disparities across prompting styles \citep{jin2026same}. This sensitivity further motivates isolating prompt components before attributing behavioral changes to a particular intermediate representation. Our setting differs by scoring a discrete intermediate memory-state label against a benchmark state mapping, rather than treating unrestricted reasoning text as the object of evaluation.

Faithfulness and shortcut-learning work motivates our conservative interpretation. Shortcut learning describes models exploiting easy surface regularities that need not match the intended construct \citep{geirhos2020shortcut}. Recent explanation-faithfulness evaluations similarly caution that an intermediate explanation or label can affect an answer without proving that it reflects the model's internal decision process \citep{matton2025walk}. Our forced-state diagnostic is therefore framed as sensitivity to supplied benchmark-associated labels, not as evidence that generated states are faithful mechanisms.

\section{Task Formulation}
\label{sec:task}

Let $H=\{h_1,\ldots,h_n\}$ be a retrieved user history and $T$ the current task. A deployed assistant ultimately produces a response $R$, but this paper evaluates only an intermediate policy $\pi$. If response generation were evaluated, a policy-conditioned decomposition would be
\[
P(R \mid H,T)=\sum_{\pi} P(\pi \mid H,T)P(R \mid H,T,\pi).
\]
Our experiments stop at estimating $\pi$.

For state-structured prompting, the model may also emit a diagnostic state $s$:
\[
s \sim P(s \mid H,T), \qquad \pi \sim P(\pi \mid s,H,T).
\]
The benchmark-implied states are Active, Stale-or-Irrelevant, Conflicting-or-Superseded, and Underspecified. In the frozen controlled set these four labels map deterministically to the four reference policies, so they should be read as benchmark-associated state labels rather than an independent latent mechanism. The policy labels \textsc{Use}, \textsc{Ignore}, \textsc{Update}, and \textsc{Ask} are a benchmark-specific routing abstraction, not a complete, universal, or mutually exhaustive ontology for deployed memory systems. We distinguish generation categories, benchmark-implied states, and reference policies throughout.

\section{Audit Protocol and Prompt Arms}
\label{sec:framework}

\begin{figure}[H]
  \centering
  \small
  \definecolor{auditPrimary}{RGB}{235,241,247}
  \definecolor{auditExplore}{RGB}{248,244,232}
  \definecolor{auditNeutral}{RGB}{242,242,242}
  \definecolor{auditBorder}{RGB}{84,89,96}
  \definecolor{auditMuted}{RGB}{92,98,105}
  \begin{tikzpicture}[
    stage/.style={
      rectangle,
      rounded corners=3pt,
      draw=auditBorder,
      line width=0.4pt,
      minimum width=0.172\textwidth,
      minimum height=2.75cm,
      text width=0.142\textwidth,
      inner xsep=7pt,
      inner ysep=5pt,
      align=center,
      font=\small
    },
    badge/.style={
      circle,
      draw=auditBorder,
      fill=white,
      line width=0.4pt,
      minimum size=1.2em,
      inner sep=0pt,
      font=\footnotesize\bfseries
    },
    flow/.style={
      -{Stealth[length=4pt,width=5pt]},
      draw=auditBorder,
      line width=0.4pt
    },
    note/.style={
      rectangle,
      rounded corners=3pt,
      draw=auditBorder,
      line width=0.4pt,
      fill=white,
      text width=0.17\textwidth,
      align=center,
      inner sep=4pt,
      minimum height=1.55cm,
      font=\footnotesize
    },
    snote/.style={
      rectangle,
      rounded corners=3pt,
      draw=auditBorder,
      line width=0.4pt,
      fill=white,
      minimum width=0.172\textwidth,
      text width=0.142\textwidth,
      inner xsep=7pt,
      inner ysep=5pt,
      align=center,
      minimum height=2.7cm,
      font=\footnotesize
    }
  ]
  \node[stage, fill=auditPrimary] (s1) {
    \hspace{0.35em}\textbf{Dataset Audit}\strut\\[0.25em]
    {\footnotesize class coverage\\lexical baselines\\label leakage}
  };
  \node[stage, fill=auditPrimary, right=0.22cm of s1] (s2) {
    \hspace{0.35em}\textbf{Prompt Isolation}\strut\\[0.25em]
    {\footnotesize taxonomy\\state output\\rationale/order}
  };
  \node[stage, fill=auditPrimary, right=0.22cm of s2] (s3) {
    \hspace{0.35em}\textbf{Label Check}\strut\\[0.25em]
    {\footnotesize answer-associated\\state labels\\diagnostic only}
  };
  \node[stage, fill=auditExplore, right=0.22cm of s3] (s4) {
    \hspace{0.35em}\textbf{Evidence Follow-Up}\strut\\[0.25em]
    {\footnotesize decomposed fields\\matched variants\\exploratory}
  };
  \node[stage, fill=auditPrimary, right=0.22cm of s4] (s5) {
    \hspace{0.35em}\textbf{Execution Audit}\strut\\[0.25em]
    {\footnotesize parse failures\\provider errors\\cost/rebuild}
  };
  
  \foreach \stageNode/\num in {s1/1,s2/2,s3/3,s4/4,s5/5} {
    \node[badge] at ($(\stageNode.north west)+(0.23cm,-0.23cm)$) {\num};
  }
  
  \draw[flow] (s1.east) -- (s2.west);
  \draw[flow] (s2.east) -- (s3.west);
  \draw[flow] (s3.east) -- (s4.west);
  \draw[flow] (s4.east) -- (s5.west);
  
  \draw[decorate, decoration={brace, mirror, amplitude=4pt}, draw=auditBorder, line width=0.4pt]
    ($(s1.south west)+(0,-0.18cm)$) -- ($(s3.south east)+(0,-0.18cm)$);
  \coordinate (primaryAnchor) at ($(s1.south)!0.5!(s3.south)$);
  \node[note, text width=0.34\textwidth, below=0.42cm of primaryAnchor] (primary) {
      \textbf{Primary evidence.} Frozen controlled set matched prompt isolation and family-cluster analysis.
  };
  \draw[decorate, decoration={brace, mirror, amplitude=4pt}, draw=auditBorder, line width=0.4pt]
    ($(s4.south west)+(0,-0.18cm)$) -- ($(s4.south east)+(0,-0.18cm)$);
  \node[snote, below=0.42cm of s4.south] (explore) {
      \textbf{Exploratory evidence.} Semantic-evidence follow-up avoids the old state label but has residual template diagnostics.
  };
  \node[snote, fill=auditNeutral, draw=auditMuted, text=auditMuted, dashed,
        below=0.42cm of s5.south] (scope) {
    \textbf{Out of scope.} Downstream responses, tool actions, and memory mutation are not evaluated.
  };
  \end{tikzpicture}
  \caption{Five-stage audit protocol instantiated in this paper. The label-conditioning diagnostic is separated from semantic-evidence interventions because supplied benchmark-state labels map deterministically to policy labels and therefore do not test mechanism faithfulness.}
  \label{fig:experiment-design}
  \end{figure}
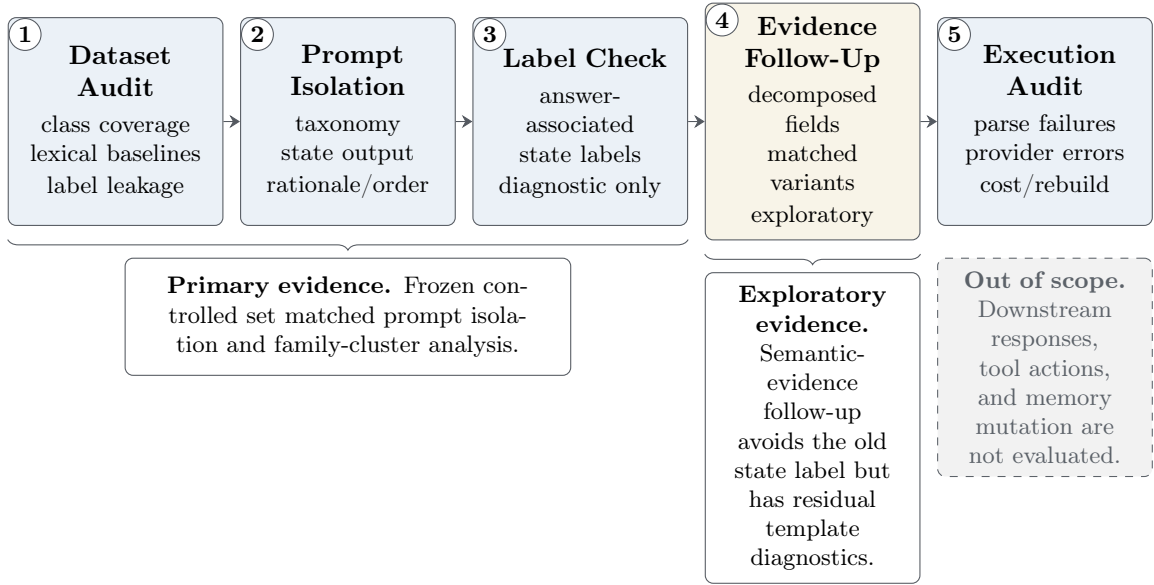

We organize the study as a five-stage audit protocol for explicit intermediate outputs in agent decisions. Stage 1 is a dataset-construction audit: check class coverage, label distribution, lexical separability, template predictability, direct label leakage, grouped generalization, and transfer limitations before interpreting prompt gains. Stage 2 is matched prompt isolation: vary one component at a time, such as taxonomy exposure, explicit state output, rationale output, output ordering, or deterministic routing. Stage 3 is an answer-associated label audit: if an intermediate label is deterministically mapped to the target answer, forced-label conditions are reported as label-conditioning diagnostics rather than mechanism evidence. Stage 4 is a semantic-evidence follow-up: replace answer-associated labels with decomposed evidence fields such as relevance, temporal status, supersession, context sufficiency, and risk or confirmation. Stage 5 is an operational audit: retain parse failures, provider request-validation failures, retries, terminal call outcomes, endpoint amendments, token/cost accounting, and release-package rebuild checks. Broader provenance work treats typed execution and evidence traces as the substrate for verification, debugging, audit, and recovery in LLM agents \citep{wang2026agent}; our retained records are a narrower provider-output accounting layer for this benchmark. Figure~\ref{fig:experiment-design} gives the audit-protocol diagram.

All evaluated arms are single-call prompts that receive the history and task and return strict JSON. Historical internal names are retained only as artifact identifiers. The development-set arms include implicit-state/no-state-output prompts, a state-structured prompt bundle, and policy-only baselines. The old state-structured bundle changes several factors at once: state taxonomy, output schema, step wording, rationale requirements, prompt length, and policy descriptions. Its gains therefore cannot be attributed to the state field alone.

The frozen controlled set matched ablation uses five pre-specified arms frozen before the corresponding hosted evaluation: policy-only, taxonomy-only, state-output, state-output-with-rationale, and deterministic-routing with policy definitions. Policy definitions and formatting are held constant where possible. The primary matched comparison is state-output versus policy-only. The deterministic-routing arm asks only for a state and maps emitted states to policies in code. Exact artifact IDs and the state-to-policy mapping for these reader-facing names are listed in Appendix~\ref{sec:app-artifact-name-mapping}.

The label-conditioning diagnostic uses six artifact conditions; Appendix~\ref{sec:realm-extended-related} describes them and Appendix~\ref{sec:app-label-conditioning-wrong-state} gives the wrong-state construction.

\section{Data}
\label{sec:benchmark}

\textbf{Development set.} MemoryPolicy-Bench contains 480 synthetic examples generated with Gemini 1.5 Flash. It is balanced across Stable, Contextual, Updated, and Ambiguous generation categories, but the reference-policy distribution is 240 \textsc{Use}, 120 \textsc{Update}, 120 \textsc{Ask}, and 0 \textsc{Ignore}. Labels are generator-assigned reference policies. The plain Always-USE majority baseline reaches 50.0\%, and a supervised TF-IDF diagnostic reaches 89.6$\pm$3.5\% policy accuracy under random folds. This construct-validity warning means the set is lexically separable and used here as development evidence only. Appendix~\ref{sec:app-historical-development-results} reports the historical development-set prompt-bundle result and why it is treated as motivation only.

\textbf{Frozen controlled set.} We construct a separate frozen counterfactual set with 160 examples: 40 scenario families crossed with \textsc{Use}, \textsc{Ignore}, \textsc{Update}, and \textsc{Ask}. Reference policies are derived deterministically from structured semantic slots before natural-language rendering. Every example has exactly four history records: an earlier temporal record, a later temporal record, and two context-specific records. The task text is byte-identical across the four variants in each family; only the semantic relation between history and task changes; Appendix~\ref{sec:moved-main-materials} gives the representative four-way family.

The controlled set is synthetically generated and rule-labeled. It passes deterministic structural and semantic validation, including policy balance, four records per example, identical within-family task text, type compatibility, and no direct label leakage. Under grouped diagnostics, the frozen controlled set remains harder for the evaluated lexical baselines: word TF-IDF macro F1 is 0.390 and character n-gram macro F1 is 0.368 in the controlled-set audit. This should not be read as a strict magnitude comparison to the development-set random-fold diagnostic, because the development set has no positive \textsc{Ignore} reference class and differs in split structure, label space, and priors. Appendix~\ref{sec:app-shortcut-transfer} reports a development-set-to-controlled-set TF-IDF transfer diagnostic. Each of the 160 examples was verified by two annotators, working separately and blind to the rule-derived reference policies; the annotators agreed with each other and with the reference policy on every example. Labels remain rule-derived and may not capture every reasonable human judgment.

\section{Experimental Setup}
\label{sec:experimental_setup}

The development-set evidence uses Groq-hosted Llama-3.3-70B (\texttt{llama-3.3-70b-versatile}) and Qwen3-32B (\texttt{qwen/qwen3-32b}). The frozen controlled set experiments use Llama-3.3-70B and GPT-OSS-120B (\texttt{openai/gpt-oss-120b}). Qwen3-32B was replaced before any controlled-set hosted result existed because the endpoint became unavailable on the current Groq tier; GPT-OSS results are never labeled as Qwen results. Llama is the bridge endpoint across both phases.

The Llama endpoint is documented by the Llama 3 family report and the Llama-3.3-70B-Instruct model card \citep{dubey2024llama3,metallama2024llama33modelcard}. The development Qwen endpoint is documented by the Qwen3 technical report \citep{yang2025qwen3}. The GPT-OSS endpoint is documented by the official GPT-OSS model card \citep{openai2025gptossmodelcard}.

All hosted controlled-set runs use the frozen 160 examples, seeds 101/102/103, temperature 0.7, hash-recorded frozen prompt templates, strict JSON parsing, raw response logging, finish-reason logging, token accounting, and resumable call IDs. Parse failures remain visible and count as incorrect. The matched ablation contains 4,800 calls; the label-conditioning diagnostic contains 5,760 calls. Frozen-core hosted cost including two smoke calls is estimated at \$1.8007 under a \$20.00 hard budget.

Primary uncertainty uses scenario-family clustering. We report paired accuracy differences, 10,000-resample percentile bootstrap 95\% intervals, and paired family-level sign-flip permutation tests with plus-one correction. Non-significant p-values are not interpreted as equivalence.

\section{Results}
\label{sec:main_results}

Policy-only denotes distinct baselines in the matched ablation, label-conditioning diagnostic, and semantic-evidence follow-up. These protocols use different prompt templates; the development comparison also changes label space and priors, while semantic-evidence uses a different rendering/template with higher lexical separability despite retaining balanced policies. Their accuracies are therefore not directly comparable; the label-conditioning table note also separates the two frozen protocol sources, and the semantic-evidence policy-only value is reported only for Llama.

Tables~\ref{tab:frozen-v3-matched-primary} and~\ref{tab:frozen-v3-label-conditioning} summarize the frozen controlled set results. The matched state-output ablation does not support a universal prompt recipe. State-output improves over clean policy-only by only 0.6 percentage points for Llama, with a 95\% family-cluster interval of [-2.9, 4.0] and p=0.8084. GPT-OSS shows a larger positive difference of 3.3 points, but the interval includes zero and the paired permutation test is not significant (p=0.0710; Holm-adjusted p=0.1420). We therefore do not claim that merely adding an explicit state output consistently improves policy accuracy.

Taxonomy-only shows a different effect: exposing the four state definitions without requiring a state output improves accuracy on both endpoints (Llama 44.0\% to 53.1\%, +9.17 points, 95\% CI [5.00, 13.33], raw p=0.0003, Holm-adjusted p=0.0012; GPT-OSS 54.8\% to 59.8\%, +5.00 points, 95\% CI [1.67, 8.33], raw p=0.0045, Holm-adjusted p=0.0135). This dissociates taxonomy exposure from state emission, but because the four benchmark states map one-to-one onto the four policies, taxonomy exposure may convey label structure rather than useful typed knowledge.

The label-conditioning diagnostic shows sensitivity to externally supplied benchmark-associated state labels. Supplying the benchmark-implied label improves routing by 7.3 points for Llama and 11.9 points for GPT-OSS, with cluster-aware intervals excluding zero. Correct and conflicting supplied labels produce routing gaps of 15.2 and 15.0 points for Llama and GPT-OSS, respectively, relative to forced-correct routing. Relative to policy-only (label-conditioning protocol) rather than forced-correct routing, forced-wrong-state accuracy is 7.9 points lower for Llama (95\% CI [-11.9, -4.2], $p<.001$) and 3.1 points lower for GPT-OSS (95\% CI [-7.1, 0.8], p=0.1582). Because the four benchmark states deterministically map to the four policies, these contrasts do not establish mechanism faithfulness.

\begin{table*}[t]
\centering
\small
\setlength{\tabcolsep}{4pt}
\resizebox{\textwidth}{!}{%
\begin{tabular}{lccp{0.22\textwidth}cp{0.22\textwidth}}
\toprule
Model & Clean acc. & Taxonomy acc. & Taxonomy $\Delta$ vs clean & State acc. & State $\Delta$ vs clean \\
\midrule
Llama-3.3-70B & 44.0\% & 53.1\% & +9.17 pp\newline [5.00, 13.33]\newline raw p=0.0003; Holm p=0.0012 & 44.6\% & 0.6 pp\newline [-2.9, 4.0]\newline raw p=0.8084; Holm p=0.8084 \\
GPT-OSS-120B & 54.8\% & 59.8\% & +5.00 pp\newline [1.67, 8.33]\newline raw p=0.0045; Holm p=0.0135 & 58.1\% & 3.3 pp\newline [0.0, 6.7]\newline raw p=0.0710; Holm p=0.1420 \\
\bottomrule
\end{tabular}%
}
\caption{Primary matched ablation on the frozen controlled set. Accuracies are exact-match policy accuracies. $\Delta$ is each arm minus clean policy-only in percentage points; parse failures count as incorrect. Parse failures are 0 for clean policy-only and state-output on both endpoints; taxonomy-only has 1 parse failure for Llama and 0 for GPT-OSS. Intervals are 95\% scenario-family-cluster bootstrap intervals; p-values are paired family-level permutation tests. Holm adjustment is over the four arm-versus-clean contrasts shown for taxonomy-only and state-output across the two endpoints; the state-output Holm values are unchanged under this expanded family. All reported percentage-point differences in the paper are computed from unrounded accuracies; displayed rounded endpoints may not reproduce them exactly.}
\label{tab:frozen-v3-matched-primary}
\end{table*}

\begin{table*}[t]
\centering
\small
\setlength{\tabcolsep}{5pt}
\resizebox{\textwidth}{!}{%
\begin{tabular}{lcccc}
\toprule
Model & Policy-only & Supplied ref. & Supplied conflict & State-first \\
\midrule
Llama-3.3-70B & 45.4\% & 52.7\% & 37.5\% & 42.7\% \\
GPT-OSS-120B & 53.1\% & 65.0\% & 50.0\% & 59.2\% \\
\midrule
Model & Ref. vs policy & Conflict vs ref. & Conflict vs policy & State-first ($\Delta$) \\
\midrule
Llama-3.3-70B & \begin{tabular}[c]{@{}c@{}}7.3 pp\\{}[2.5, 12.1]\\p=0.0065\end{tabular} & \begin{tabular}[c]{@{}c@{}}-15.2 pp\\{}[-20.4, -10.0]\\\mbox{$p<.001$}\end{tabular} & \begin{tabular}[c]{@{}c@{}}-7.9 pp\\{}[-11.9, -4.2]\\\mbox{$p<.001$}\end{tabular} & \begin{tabular}[c]{@{}c@{}}-2.7 pp\\p=0.0825\end{tabular} \\
GPT-OSS-120B & \begin{tabular}[c]{@{}c@{}}11.9 pp\\{}[8.3, 15.4]\\\mbox{$p<.001$}\end{tabular} & \begin{tabular}[c]{@{}c@{}}-15.0 pp\\{}[-20.0, -10.2]\\\mbox{$p<.001$}\end{tabular} & \begin{tabular}[c]{@{}c@{}}-3.1 pp\\{}[-7.1, 0.8]\\p=0.1582\end{tabular} & \begin{tabular}[c]{@{}c@{}}6.0 pp\\p=0.0025\end{tabular} \\
\bottomrule
\end{tabular}%
}
\caption{Label-conditioning sensitivity diagnostic on the frozen controlled set. The supplied-reference and supplied-conflict columns are absolute accuracies when the prompt receives a benchmark-associated state label. Delta columns are percentage-point differences on matched controlled-set decisions; conflict-vs-reference is relative to supplied-reference routing, not policy-only. Because controlled-set state labels map deterministically to policies, this diagnostic tests sensitivity to supplied benchmark-associated labels rather than an internal causal state mechanism.}
\label{tab:frozen-v3-label-conditioning}
\end{table*}

\FloatBarrier

Natural state elicitation remains model-dependent. State-first is negative or neutral for Llama relative to policy-only (-2.7 points, p=0.0825), but positive for GPT-OSS (+6.0 points, p=0.0025). Benchmark-mapped state agreement in this condition is 41.5\% for Llama and 55.8\% for GPT-OSS. In the state-only deterministic-routing condition (label-conditioning protocol), where policy is computed from the emitted state, policy accuracy is 46.9\% for Llama and 52.5\% for GPT-OSS. This pattern supports the narrower conclusion that explicit state elicitation is not enough: accurately predicting the benchmark-defined state remains difficult for the evaluated models. The old development-set prompt-bundle gain cannot be reduced to the state output field because the bundle changed multiple prompt dimensions and was measured on a lexically separable development set.

The semantic-evidence follow-up protocol replaces the old four-state label with decomposed evidence fields for task relevance, temporal evidence, supersession, context sufficiency, and risk or confirmation. Explicit semantic-evidence output reduces accuracy relative to policy-only by 7.5 percentage points (95\% CI [-12.7, -2.5], Holm-adjusted p=0.0146 within the three arm comparisons in this follow-up protocol), while explicit output differs little from taxonomy-only prompting (+1.3 points, p=0.6079). The taxonomy-only arm itself is lower than policy-only by 8.8 points (95\% CI [-12.1, -5.6], Holm-adjusted p=0.0003). Thus, for the cleanly evaluated endpoint, degradation is more consistent with taxonomy exposure or the added structured decision frame than with a benefit from mandatory intermediate serialization. For GPT-OSS, the explicit-evidence arm produced provider-side JSON validation failures for all attempted records, so that endpoint does not yield a clean behavioral comparison for this arm. We therefore treat semantic-evidence elicitation as negative for Llama and unavailable for GPT-OSS, not as evidence that decomposed intermediate fields solve the leakage concern. Appendix~\ref{sec:app-semantic-evidence-followup} gives the full follow-up protocol, design audit, and per-arm accuracies.

\subsection{Counterfactual Consistency and Failure Analysis}

The four-way family structure exposes failures that average example accuracy hides. In the primary matched ablation, Llama solves no complete four-variant family in either policy-only or state-output form; GPT-OSS solves 1.7\% of family-seed units in policy-only form and 5.0\% with state output. Most families have only two of four variants correct: under policy-only prompting, Llama has 2/4 correct variants in 65.0\% of family-seed units and never reaches 4/4.

Seed stability and error taxonomy further localize failures. In the primary matched ablation, Llama policy-only predictions are unanimous across seeds on 69.4\% of examples, while state-output predictions are unanimous on 76.2\%; GPT-OSS shows 83.1\% and 74.4\%. Among memory-reasoning error modes, the most frequent by count are conflict over-detection, task neglect (the model ignores the current task text), and recency overreach, followed by ask-defaulting and irrelevance-to-\textsc{Ask} confusion (see Appendix~\ref{sec:app-error-taxonomy}). These are non-exclusive rule-based diagnostics from metadata and predictions, not human-coded annotations, and their counts must not be summed.

\section{Discussion}
\label{sec:discussion}

\textbf{Supplied benchmark-associated labels are not self-eliciting states.} The forced-state artifact conditions show that policy routing is sensitive to externally supplied benchmark-associated labels: when the benchmark-implied label is supplied by the experimenter, routing improves, and a conflicting supplied label shifts predictions away from the reference mapping. The matched prompt ablation shows a different fact: asking the model to emit a state field does not consistently produce the needed improvement. These results are compatible because accurately predicting the benchmark-defined state remains difficult for the evaluated models. The rendered taxonomy prompt lists the four states in the same order as the four policies and closely paraphrases them; the mapping is recoverable without being stated, so the taxonomy-only gain is confounded with definition restatement. For example, Conflicting-or-Superseded mirrors Update's newer-overrides-older wording, and Stale-or-Irrelevant explicitly says ``no clarification is needed,'' separating Ignore from Ask.

\textbf{Conflicting supplied labels can shift policy predictions.} The forced-wrong-state condition creates an approximately 15-point routing gap relative to the forced-correct-state condition for both endpoints. Relative to policy-only, it is lower by 7.9 points for Llama and 3.1 points for GPT-OSS. This cautions against treating explicit benchmark-state outputs as an automatic auditing guarantee. Externally supplied benchmark-associated labels can shift policy predictions, including when the supplied label conflicts with the reference mapping; systems need state-quality checks, abstention, or clarification mechanisms rather than state fields alone.

\textbf{Model dependence matters.} GPT-OSS benefits from natural state-first prompting on the frozen controlled set, while Llama does not. Llama also performs better with policy-first ordering than state-first ordering in the diagnostic table. This makes a universal ``state first'' prompting recommendation scientifically too strong. The per-policy decomposition makes this concrete: the state field pushes each model toward a different policy and away from another, and even flips the sign of its effect on Ask across models, so its effect is a model-specific bias shift rather than a general gain.

\textbf{Family-level analysis changes the error picture.} Example-level accuracy makes the task look like a partially solved four-class classification problem, but family-level exact match (Table~\ref{tab:app-family-counterfactual}) shows that models rarely solve all four matched counterfactual variants in the same scenario. Pairwise transition analysis further separates a correct policy change from merely changing to another wrong label. This is why matched counterfactual families are useful as an audit unit: they reveal whether a prompt tracks the intended semantic relation, rather than merely improving marginal accuracy.

\textbf{State output redistributes policies rather than improving them.} The overall matched-ablation null hides opposing class-level effects that cancel. Adding a state field significantly shifts Llama toward Use (+17.5 pp, 95\% CI [6.67, 28.33], Holm $p$=0.010, within-model four-policy family) and away from Ask (-14.2 pp, 95\% CI [-22.50, -6.67], Holm $p$=0.010), while it shifts GPT-OSS toward Ignore (+15.8 pp, 95\% CI [7.50, 25.00], Holm $p$=0.003) and away from Update (-8.3 pp, 95\% CI [-13.33, -3.33], Holm $p$=0.010). The gains and losses are of similar magnitude and net to the near-zero overall change. The Ask policy makes the mechanism concrete and shows it is model-specific in sign: with a state field, Llama's Ask output nearly collapses (precision/recall/F1 fall from 35.3/15.0/21.1 to 2.9/0.8/1.3), whereas GPT-OSS's Ask precision/recall rises (24.53/10.83 to 44.19/15.83). The same intervention therefore suppresses a policy in one model and sharpens it in another, so it is a model-specific bias shift rather than a general gain --- which is why family-level exact match stays low and why a universal ``state-first'' recipe is unwarranted.

\textbf{The bottleneck is the decision mapping, not information availability.} We also tested the opposite intervention: instead of eliciting structure on the output side, we normalized the input, replacing the natural-language history with explicit [subject|attribute|value|tag] tuples derived from the same structured slots (a probe that does not answer-leak: grouped lexical decoders reach only 0.375 word / 0.33125 character accuracy). This did not help and significantly hurt: normalized input lowered policy accuracy by 8.5 points for Llama-3.3-70B (95\% CI [-13.96, -3.54], p=0.0034) and 7.9 points for GPT-OSS-120B (95\% CI [-12.08, -3.54], p=0.0012). Together with the null output-side ablation and the imperfect forced-correct-state routing, this is consistent with a decision-mapping bottleneck: making relations explicitly available on the input side did not help. Appendix~\ref{sec:app-normalized-input-protocol} gives the protocol and rendered prompt.

\section{Conclusion}

The strongest supported conclusion is not that explicit state output reliably improves memory-policy classification. On a frozen four-policy controlled counterfactual set, the matched state-output ablation is null for Llama and marginal for GPT-OSS. A separate label-conditioning diagnostic shows that benchmark-associated supplied labels shift routing, and correct versus conflicting supplied labels create an approximately 15-point routing gap relative to the forced-correct condition. Because these labels deterministically map to policies, this is not evidence of a faithful internal state mechanism. Explicit state elicitation requires more than adding a field to the JSON schema.

This reframes MemoryPolicy-Bench as an audit of memory-policy decisions and of the intermediate fields used to justify them. Development-set prompt-bundle gains are useful signals, but the controlled counterfactual evidence shows that future progress should target robust semantic-evidence extraction, calibrated asking, provider-compatible structured-output protocols, and downstream response/action evaluation rather than relying on a generic state-first prompt recipe. More broadly, the five-stage audit protocol is proposed and instantiated here on memory-policy classification; within this task family, it provides a reusable way to separate dataset shortcuts, bundled prompt changes, answer-associated labels, stochastic instability, and provider failures before making claims about structured intermediate outputs.

\section*{Limitations}

Both datasets are synthetic. The frozen controlled set is controlled and counterfactual rather than naturalistic conversation data. Its labels are rule-derived from structured semantics, deterministically validated, and verified by two annotators (blind to the reference labels) who agreed with each other and with the reference policies on all 160 examples. Rule-derived labels encode the benchmark designers' intended policy abstraction and may not capture every reasonable human judgment.

The endpoint panel differs across phases. The development set uses Llama-3.3-70B and Qwen3-32B; the frozen controlled set uses Llama-3.3-70B and GPT-OSS-120B after the Qwen endpoint became unavailable before any controlled-set hosted result existed. This weakens direct two-model replication claims across phases. A third endpoint (llama-3.1-8b-instant) reproduces the direction of the state-output effect in Appendix~\ref{sec:app-third-endpoint-replication}, but with elevated parse failures, so cross-model state-output behavior remains model-dependent.

The experiments evaluate policy classification only. They do not measure downstream response quality, whether old memory is actually applied or suppressed, tool-action success, retrieval quality, or memory-store mutation. Recent operational systems address transactional memory commits, provenance-aware shared memory, rollback repair from faulty memories, and proactive recovery from multi-step execution failures \citep{li2026memtx,wang2026mapgraph,yu2026faulty,xu2026reaction}; those settings mutate persistent state or execute actions, whereas our benchmark stops before that layer. The forced-state artifact conditions also idealize state availability and are best interpreted as label-conditioning sensitivity diagnostics: because benchmark states map deterministically to policies, they do not guarantee that a deployed model will infer or use a faithful internal state from natural text.

The four-policy taxonomy is intentionally narrow. Real systems may need multiple actions at once: an update may also require using the new value, ignoring one memory may still require asking about another field, and high-risk tasks may require confirmation even when a memory appears active. Deletion, forgetting, privacy restrictions, scope constraints, and multi-turn repair are outside the current benchmark. Exact-match single-label scoring therefore does not cover the full strategy space of deployed memory agents. Independent third-party annotation would be useful future validation, but is not claimed in the current paper.

Finally, hosted API behavior and model endpoints may evolve. The local run records preserve raw responses, finish reasons, token counts, parse failures, call IDs, and cost logs so that later runs can be audited against the frozen protocol.
The semantic-evidence follow-up further illustrates this operational limitation: GPT-OSS returned provider-side JSON validation failures for the explicit-evidence arm under the frozen request format. We retain those records in execution accounting rather than changing the prompt after seeing outcomes, but we do not interpret that endpoint-arm failure as model policy behavior.

\section*{Ethics Statement}
This work studies long-term user memory, a setting that can create privacy and autonomy risks if deployed carelessly. MemoryPolicy-Bench is synthetic and does not contain real user logs, private conversations, or personally identifiable information. This reduces immediate data exposure, but it also means our conclusions should be validated before deployment in systems that store real user histories.

The proposed framework is intended to reduce inappropriate memory use by making \textit{Ignore} and \textit{Ask} explicit policy options. However, state classification itself requires processing potentially sensitive history. Practical systems should minimize retained data, support user inspection and deletion of memories, and avoid applying personal information outside the context for which it was provided. Differential privacy and other user-level protections \citep{mcmahan2018learning} are complementary to this policy-layer approach.

There is also a user-experience risk. Agents that over-ask may frustrate users or shift cognitive labor onto them, while agents that under-ask may apply stale or irrelevant preferences. More autonomous downstream agents add further interaction surfaces: adjacent Text-to-SQL and embodied-control work studies iterative refinement, multi-agent collaboration, and long-horizon action failures, whereas our audit stops before response generation, tool execution, or environment mutation \citep{su2026agentic,liu2026palm}. Our results suggest that calibrated clarification should be evaluated as a precision-recall problem, not as a simple preference for more questions. Documentation of the synthetic generation process and intended use should accompany any future public code or data release.

\bibliographystyle{plainnat}
\bibliography{references}

@article{packer2023memgpt,
  title = {{MemGPT}: Towards {LLM}s as Operating Systems},
  author = {Packer, Charles and Wooders, Sarah and Lin, Kevin and Fang, Vivian and Patil, Shishir G. and Stoica, Ion and Gonzalez, Joseph E.},
  journal = {arXiv preprint arXiv:2310.08560},
  year = {2023},
  url = {https://arxiv.org/abs/2310.08560}
}

@inproceedings{xu2025amem,
  title = {{A-MEM}: Agentic Memory for {LLM} Agents},
  author = {Xu, Wujiang and Liang, Zujie and Mei, Kai and Gao, Hang and Tan, Juntao and Zhang, Yongfeng},
  booktitle = {Advances in Neural Information Processing Systems},
  year = {2025},
  url = {https://proceedings.neurips.cc/paper_files/paper/2025/hash/19909c36f51abc4856b4560aff3d36d6-Abstract-Conference.html}
}

@inproceedings{wu2024longmemeval,
  title = {{LongMemEval}: Benchmarking Chat Assistants on Long-Term Interactive Memory},
  author = {Wu, Di and Wang, Hongwei and Yu, Wenhao and Zhang, Yuwei and Chang, Kai-Wei and Yu, Dong},
  booktitle = {International Conference on Learning Representations},
  year = {2025},
  url = {https://proceedings.iclr.cc/paper_files/paper/2025/hash/d813d324dbf0598bbdc9c8e79740ed01-Abstract-Conference.html}
}

@inproceedings{salemi2024lamp,
  title = {{LaMP}: When Large Language Models Meet Personalization},
  author = {Salemi, Alireza and Mysore, Sheshera and Bendersky, Michael and Zamani, Hamed},
  booktitle = {Proceedings of the 62nd Annual Meeting of the Association for Computational Linguistics (Volume 1: Long Papers)},
  pages = {7370--7392},
  year = {2024},
  month = aug,
  address = {Bangkok, Thailand},
  publisher = {Association for Computational Linguistics},
  doi = {10.18653/v1/2024.acl-long.399},
  url = {https://aclanthology.org/2024.acl-long.399/}
}

@inproceedings{jiang2024personallm,
  title = {{PersonaLLM}: Investigating the Ability of Large Language Models to Express Personality Traits},
  author = {Jiang, Hang and Zhang, Xiajie and Cao, Xubo and Breazeal, Cynthia and Roy, Deb and Kabbara, Jad},
  booktitle = {Findings of the Association for Computational Linguistics: NAACL 2024},
  pages = {3605--3627},
  year = {2024},
  month = jun,
  address = {Mexico City, Mexico},
  publisher = {Association for Computational Linguistics},
  doi = {10.18653/v1/2024.findings-naacl.229},
  url = {https://aclanthology.org/2024.findings-naacl.229/}
}

@inproceedings{lewis2020retrieval,
  title = {Retrieval-Augmented Generation for Knowledge-Intensive {NLP} Tasks},
  author = {Lewis, Patrick and Perez, Ethan and Piktus, Aleksandra and Petroni, Fabio and Karpukhin, Vladimir and Goyal, Naman and K{\"u}ttler, Heinrich and Lewis, Mike and Yih, Wen-tau and Rockt{\"a}schel, Tim and Riedel, Sebastian and Kiela, Douwe},
  booktitle = {Advances in Neural Information Processing Systems},
  volume = {33},
  pages = {9459--9474},
  year = {2020},
  url = {https://proceedings.neurips.cc/paper/2020/hash/6b493230205f780e1bc26945df7481e5-Abstract.html}
}

@inproceedings{xu2024knowledgeconflicts,
  title = {Knowledge Conflicts for {LLM}s: A Survey},
  author = {Xu, Rongwu and Qi, Zehan and Guo, Zhijiang and Wang, Cunxiang and Wang, Hongru and Zhang, Yue and Xu, Wei},
  booktitle = {Proceedings of the 2024 Conference on Empirical Methods in Natural Language Processing},
  pages = {8541--8565},
  year = {2024},
  month = nov,
  address = {Miami, Florida, USA},
  publisher = {Association for Computational Linguistics},
  doi = {10.18653/v1/2024.emnlp-main.486},
  url = {https://aclanthology.org/2024.emnlp-main.486/}
}

@inproceedings{asai2024selfrag,
  title = {{Self-RAG}: Learning to Retrieve, Generate, and Critique through Self-Reflection},
  author = {Asai, Akari and Wu, Zeqiu and Wang, Yizhong and Sil, Avirup and Hajishirzi, Hannaneh},
  booktitle = {International Conference on Learning Representations},
  year = {2024},
  url = {https://openreview.net/forum?id=hSyW5go0v8}
}

@inproceedings{xu2019asking,
  title = {Asking Clarification Questions in Knowledge-Based Question Answering},
  author = {Xu, Jingjing and Wang, Yuechen and Tang, Duyu and Duan, Nan and Yang, Pengcheng and Zeng, Qi and Zhou, Ming and Sun, Xu},
  booktitle = {Proceedings of the 2019 Conference on Empirical Methods in Natural Language Processing and the 9th International Joint Conference on Natural Language Processing (EMNLP-IJCNLP)},
  pages = {1618--1629},
  year = {2019},
  month = nov,
  address = {Hong Kong, China},
  publisher = {Association for Computational Linguistics},
  doi = {10.18653/v1/D19-1172},
  url = {https://aclanthology.org/D19-1172/}
}

@article{kuhn2023clam,
  title = {{CLAM}: Selective Clarification for Ambiguous Questions with Generative Language Models},
  author = {Kuhn, Lorenz and Gal, Yarin and Farquhar, Sebastian},
  journal = {arXiv preprint arXiv:2212.07769},
  year = {2022},
  url = {https://arxiv.org/abs/2212.07769}
}

@inproceedings{carlini2021extracting,
  title = {Extracting Training Data from Large Language Models},
  author = {Carlini, Nicholas and Tramer, Florian and Wallace, Eric and Jagielski, Matthew and Herbert-Voss, Ariel and Lee, Katherine and Roberts, Adam and Brown, Tom B. and Song, Dawn and Erlingsson, Ulfar and Oprea, Alina and Raffel, Colin},
  booktitle = {USENIX Security Symposium},
  pages = {2633--2650},
  year = {2021},
  url = {https://www.usenix.org/conference/usenixsecurity21/presentation/carlini-extracting}
}

@inproceedings{staab2023beyond,
  title = {Beyond Memorization: Violating Privacy Via Inference with Large Language Models},
  author = {Staab, Robin and Vero, Mark and Balunovi{\'c}, Mislav and Vechev, Martin},
  booktitle = {International Conference on Learning Representations},
  year = {2024},
  url = {https://openreview.net/forum?id=kmn0BhQk7p}
}

@inproceedings{mcmahan2018learning,
  title = {Learning Differentially Private Recurrent Language Models},
  author = {McMahan, H. Brendan and Ramage, Daniel and Talwar, Kunal and Zhang, Li},
  booktitle = {International Conference on Learning Representations},
  year = {2018},
  url = {https://openreview.net/forum?id=BJ0hF1Z0b}
}

@inproceedings{maharana2024locomo,
  title = {Evaluating Very Long-Term Conversational Memory of {LLM} Agents},
  author = {Maharana, Adyasha and Lee, Dong-Ho and Tulyakov, Sergey and Bansal, Mohit and Barbieri, Francesco and Fang, Yuwei},
  booktitle = {Proceedings of the 62nd Annual Meeting of the Association for Computational Linguistics (Volume 1: Long Papers)},
  pages = {13851--13870},
  year = {2024},
  month = aug,
  address = {Bangkok, Thailand},
  publisher = {Association for Computational Linguistics},
  doi = {10.18653/v1/2024.acl-long.747},
  url = {https://aclanthology.org/2024.acl-long.747/}
}

@inproceedings{zhao2025prefeval,
  title = {Do {LLM}s Recognize Your Preferences? Evaluating Personalized Preference Following in {LLM}s},
  author = {Zhao, Siyan and Hong, Mingyi and Liu, Yang and Hazarika, Devamanyu and Lin, Kaixiang},
  booktitle = {International Conference on Learning Representations},
  year = {2025},
  url = {https://openreview.net/forum?id=QWunLKbBGF}
}

@article{dubey2024llama3,
  title = {The {Llama} 3 Herd of Models},
  author = {Grattafiori, Aaron and Dubey, Abhimanyu and Jauhri, Abhinav and Pandey, Abhinav and Kadian, Abhishek and Al-Dahle, Ahmad and Letman, Aiesha and Mathur, Akhil and Schelten, Alan and Vaughan, Alex and Yang, Amy and Fan, Angela and others},
  journal = {arXiv preprint arXiv:2407.21783},
  year = {2024},
  url = {https://arxiv.org/abs/2407.21783}
}

@misc{metallama2024llama33modelcard,
  title = {{Llama} 3.3 70B Instruct Model Card},
  author = {{Meta Llama}},
  year = {2024},
  howpublished = {\url{https://huggingface.co/meta-llama/Llama-3.3-70B-Instruct}},
  note = {Official model card; release date December 6, 2024}
}

@article{yang2025qwen3,
  title = {{Qwen3} Technical Report},
  author = {Yang, An and Li, Anfeng and Yang, Baosong and Zhang, Beichen and Hui, Binyuan and Zheng, Bo and Yu, Bowen and Gao, Chang and Huang, Chengen and Lv, Chenxu and Zheng, Chujie and Liu, Dayiheng and Zhou, Fan and Huang, Fei and Hu, Feng and Ge, Hao and Wei, Haoran and Lin, Huan and Tang, Jialong and Yang, Jian and Tu, Jianhong and Zhang, Jianwei and Yang, Jianxin and Yang, Jiaxi and Zhou, Jing and Zhou, Jingren and Lin, Junyang and Dang, Kai and Bao, Keqin and Yang, Kexin and Yu, Le and Deng, Lianghao and Li, Mei and Xue, Mingfeng and Li, Mingze and Zhang, Pei and Wang, Peng and Zhu, Qin and Men, Rui and Gao, Ruize and Liu, Shixuan and Luo, Shuang and Li, Tianhao and Tang, Tianyi and Yin, Wenbiao and Ren, Xingzhang and Wang, Xinyu and Zhang, Xinyu and Ren, Xuancheng and Fan, Yang and Su, Yang and Zhang, Yichang and Zhang, Yinger and Wan, Yu and Liu, Yuqiong and Wang, Zekun and Cui, Zeyu and Zhang, Zhenru and Zhou, Zhipeng and Qiu, Zihan},
  journal = {arXiv preprint arXiv:2505.09388},
  year = {2025},
  url = {https://arxiv.org/abs/2505.09388}
}

@misc{openai2025gptossmodelcard,
  title = {{gpt-oss-120b} and {gpt-oss-20b} Model Card},
  author = {{OpenAI}},
  year = {2025},
  howpublished = {\url{https://openai.com/index/gpt-oss-model-card/}},
  note = {Official model card, August 5, 2025}
}

@inproceedings{wei2022chain,
  title = {Chain-of-Thought Prompting Elicits Reasoning in Large Language Models},
  author = {Wei, Jason and Wang, Xuezhi and Schuurmans, Dale and Bosma, Maarten and Xia, Fei and Chi, Ed H. and Le, Quoc V. and Zhou, Denny},
  booktitle = {Advances in Neural Information Processing Systems},
  volume = {35},
  pages = {24824--24837},
  year = {2022},
  url = {https://proceedings.neurips.cc/paper_files/paper/2022/hash/9d5609613524ecf4f15af0f7b31abca4-Abstract-Conference.html}
}

@inproceedings{zhou2023least,
  title = {Least-to-Most Prompting Enables Complex Reasoning in Large Language Models},
  author = {Zhou, Denny and Scharli, Nathanael and Hou, Le and Wei, Jason and Scales, Nathan and Wang, Xuezhi and Schuurmans, Dale and Cui, Claire and Bousquet, Olivier and Le, Quoc V. and Chi, Ed H.},
  booktitle = {International Conference on Learning Representations},
  year = {2023},
  url = {https://openreview.net/forum?id=WZH7099tgfM}
}

@inproceedings{press2023measuring,
  title = {Measuring and Narrowing the Compositionality Gap in Language Models},
  author = {Press, Ofir and Zhang, Muru and Min, Sewon and Schmidt, Ludwig and Smith, Noah A. and Lewis, Mike},
  booktitle = {Findings of the Association for Computational Linguistics: EMNLP 2023},
  pages = {5687--5711},
  year = {2023},
  publisher = {Association for Computational Linguistics},
  doi = {10.18653/v1/2023.findings-emnlp.378},
  url = {https://aclanthology.org/2023.findings-emnlp.378/}
}

@article{jin2026same,
  title = {Same Question, Different Answer? Measuring and Mitigating Prompt Privilege for Equitable AI Access},
  author = {Jin, Lier and Hu, Lan and Shen, Binqi and Cai, Hanyu and Xin, Yuting},
  journal = {arXiv preprint arXiv:2608.08942},
  year = {2026},
  doi = {10.48550/arXiv.2608.08942},
  url = {https://arxiv.org/abs/2608.08942}
}

@misc{wu2026evidenceunitfairnesslimitsqueryadaptive,
  title = {Evidence-Unit Fairness and the Limits of Query-Adaptive Sparse-Dense Fusion in Financial Document Retrieval},
  author = {Wu, Chenyu and Lin, You},
  year = {2026},
  eprint = {2608.00183},
  archivePrefix = {arXiv},
  primaryClass = {cs.CE},
  url = {https://arxiv.org/abs/2608.00183}
}

@article{xin2026driftguard,
  title = {DriftGuard: Safety-Aware Multi-Monitor Detection and Selective Adaptation for Evolving Toxicity Moderation},
  author = {Xin, Yuting and Cai, Hanyu and Shen, Binqi and Jin, Lier and Hu, Lan},
  journal = {arXiv preprint arXiv:2606.28725},
  year = {2026},
  doi = {10.48550/arXiv.2606.28725},
  url = {https://arxiv.org/abs/2606.28725}
}

@misc{han2026earlyearlyenoughdesigndependent,
  title = {How Early Is Early Enough? Design-Dependent Observation-Window Sufficiency in Subscription Churn Prediction},
  author = {Han, Xiao and Xiao, Yao and Wu, Chenyu and Zhang, Tongchen},
  year = {2026},
  eprint = {2607.00473},
  archivePrefix = {arXiv},
  primaryClass = {cs.LG},
  doi = {10.48550/arXiv.2607.00473},
  url = {https://doi.org/10.48550/arXiv.2607.00473}
}

@misc{wu2026classweightingversusconditioning,
  title = {Class Weighting versus Amount Conditioning in Credit-Card Fraud Detection: A Dollar-Metric Study with a Temporal Explanation Audit},
  author = {Wu, Chenyu},
  year = {2026},
  eprint = {2607.14686},
  archivePrefix = {arXiv},
  primaryClass = {cs.CE},
  doi = {10.48550/arXiv.2607.14686},
  url = {https://doi.org/10.48550/arXiv.2607.14686}
}

@misc{han2026interpretablevslearnedencoders,
  title = {Interpretable vs Learned Encoders for High-Cardinality Fraud Detection},
  author = {Han, Xiao and Liu, Jingjing and Zheng, Moxuan and Zhang, Zhen and Wu, Chenyu},
  year = {2026},
  eprint = {2607.00477},
  archivePrefix = {arXiv},
  primaryClass = {cs.LG},
  doi = {10.48550/arXiv.2607.00477},
  url = {https://doi.org/10.48550/arXiv.2607.00477}
}

@misc{li2026prefixunderstandadaptuser,
  title = {{PrefIx}: Understand and Adapt to User Preference in Human-Agent Interaction},
  author = {Li, Jialin and Chen, Zhenhao and Luo, Hanjun and Salam, Hanan},
  year = {2026},
  eprint = {2602.06714},
  archivePrefix = {arXiv},
  primaryClass = {cs.HC},
  doi = {10.48550/arXiv.2602.06714},
  url = {https://doi.org/10.48550/arXiv.2602.06714}
}

@article{liu2026memory,
  title = {The Memory Curse: How Expanded Recall Erodes Cooperative Intent in {LLM} Agents},
  author = {Liu, Jiayuan and Li, Tianqin and Du, Shiyi and Luo, Xin and Zeng, Haoxuan and Tewolde, Emanuel and Lee, Tai Sing and Wang, Tonghan and Kingsford, Carl and Conitzer, Vincent},
  journal = {arXiv preprint arXiv:2605.08060},
  year = {2026},
  eprint = {2605.08060},
  archivePrefix = {arXiv},
  primaryClass = {cs.CL},
  doi = {10.48550/arXiv.2605.08060},
  url = {https://doi.org/10.48550/arXiv.2605.08060}
}

@misc{wu2026regdivergence,
  title = {{RegDivergence-101}: An {LLM} Benchmark for Cross-Jurisdiction Regulatory Contradiction Detection in Life Sciences},
  author = {Wu, Chuchu and Zhou, Zhiyin and Hu, Jingzhuo and You, Liang},
  year = {2026},
  note = {Preprint},
  url = {https://www.researchgate.net/doi/10.13140/RG.2.2.32171.20000}
}

@inproceedings{su2026agentic,
  title = {Agentic-{SQL} Taxonomy: The Research of Autonomous and Interactive Text-to-{SQL} with {LLMs}},
  author = {Su, Yiyun and Zhu, Huiying and Tian, Yu and Zhao, Changruo and Peng, Zujun and Liu, Yuting and Zhang, Luyan and Fan, Liang and Li, Baihua},
  editor = {Huang, De-Shuang and Zhang, Chuanlei and Chen, Wei and Li, Bo and Zhang, Qinhu and Bao, Wenzheng and Pan, Yijie and Premaratne, Prashan},
  booktitle = {Advanced Intelligent Computing Technology and Applications},
  series = {Lecture Notes in Computer Science},
  volume = {16655},
  pages = {323--334},
  year = {2026},
  publisher = {Springer Nature Singapore},
  address = {Singapore},
  isbn = {978-981-92-3438-7},
  doi = {10.1007/978-981-92-3438-7_27},
  url = {https://doi.org/10.1007/978-981-92-3438-7_27}
}

@inproceedings{liu2026palm,
  title = {{PALM}: Progress-Aware Policy Learning via Affordance Reasoning for Long-Horizon Robotic Manipulation},
  author = {Liu, Yuanzhe and Zhu, Jingyuan and Mo, Yuchen and Li, Gen and Cao, Xu and Jin, Jin and Shen, Yifan and Li, Zhengyuan and Yu, Tianjiao and Yuan, Wenzhen and Ding, Fangqiang and Lourentzou, Ismini},
  booktitle = {Proceedings of the IEEE/CVF Conference on Computer Vision and Pattern Recognition (CVPR)},
  pages = {28096--28110},
  year = {2026},
  month = jun,
  eprint = {2601.07060},
  archivePrefix = {arXiv},
  primaryClass = {cs.RO},
  url = {https://arxiv.org/abs/2601.07060}
}

@inproceedings{Fu_2026,
  title = {{S-Path-RAG}: Semantic-Aware Shortest-Path Retrieval Augmented Generation for Multi-Hop Knowledge Graph Question Answering},
  author = {Fu, Rong and Wang, Yemin and Xu, Tianxiang and Liu, Yongtai and Tang, Weizhi and Wu, Wangyu and Ma, Xiaowen and Fong, Simon},
  booktitle = {Proceedings of the ACM Web Conference 2026},
  pages = {4057--4068},
  year = {2026},
  month = apr,
  publisher = {ACM},
  doi = {10.1145/3774904.3792459},
  url = {https://doi.org/10.1145/3774904.3792459},
  eprint = {2603.23512},
  archivePrefix = {arXiv},
  primaryClass = {cs.CL}
}

@misc{wang2026agent,
  title = {From Agent Traces to Trust: A Survey of Evidence Tracing and Execution Provenance in {LLM} Agents},
  author = {Wang, Yiqi and Zhang, Jiaqi and Cai, Taotao and Liu, Zirui and Sun, Qingqiang and Sun, Zequn and Wu, Zhangkai and Dong, Manqing and Zheng, Mingkai and Yin, Xuefei and Zhu, Yanming},
  year = {2026},
  eprint = {2606.04990},
  archivePrefix = {arXiv},
  primaryClass = {cs.CR},
  doi = {10.48550/arXiv.2606.04990},
  url = {https://arxiv.org/abs/2606.04990}
}

@misc{li2026memtx,
  title = {{MemTX}: Transactional Belief Commit for Stateful Agent Memory},
  author = {Li, Xiaoyang and Wang, Yiqi and Lu, Haohui and Chen, Zhi and Li, Mo and Song, Pingan and Zheng, Mingkai and Cai, Taotao},
  year = {2026},
  eprint = {2607.23929},
  archivePrefix = {arXiv},
  primaryClass = {cs.AI},
  doi = {10.48550/arXiv.2607.23929},
  url = {https://arxiv.org/abs/2607.23929}
}

@misc{yu2026faulty,
  title = {From Faulty Memories to Corrected Actions: Dependency-Guided Rollback Repair for Memory-Augmented Agents},
  author = {Yu, Caili and Wang, Yiqi and Zhang, Jiaqi and Duan, Yiqun and Zheng, Mingkai and Wu, Zhangkai and Shi, Kaize and Cai, Taotao},
  year = {2026},
  eprint = {2608.10502},
  archivePrefix = {arXiv},
  primaryClass = {cs.AI},
  doi = {10.48550/arXiv.2608.10502},
  url = {https://arxiv.org/abs/2608.10502}
}

@misc{wang2026mapgraph,
  title = {{MAP-Graph}: Provenance-Aware Shared Memory for Multi-Agent Workflows},
  author = {Wang, Yiqi and Yan, Zihao and Zhang, Jiaqi and Wu, Zhangkai and Zheng, Mingkai and Sun, Zequn and Zhu, Yanming and Cai, Taotao},
  year = {2026},
  eprint = {2608.10509},
  archivePrefix = {arXiv},
  primaryClass = {cs.AI},
  doi = {10.48550/arXiv.2608.10509},
  url = {https://arxiv.org/abs/2608.10509}
}

@misc{xu2026reaction,
  title = {From Reaction to Anticipation: Proactive Failure Recovery through Agentic Task Graph for Robotic Manipulation},
  author = {Xu, Sheng and Jin, Ruixing and Zhou, Huayi and Yue, Bo and Qiao, Guanren and Tai, Yunxin and Deng, Yueci and Jia, Kui and Liu, Guiliang},
  year = {2026},
  eprint = {2605.11951},
  archivePrefix = {arXiv},
  primaryClass = {cs.RO},
  doi = {10.48550/arXiv.2605.11951},
  url = {https://arxiv.org/abs/2605.11951},
  note = {Accepted to RSS 2026}
}

@article{geirhos2020shortcut,
  title = {Shortcut Learning in Deep Neural Networks},
  author = {Geirhos, Robert and Jacobsen, J{\"o}rn-Henrik and Michaelis, Claudio and Zemel, Richard and Brendel, Wieland and Bethge, Matthias and Wichmann, Felix A.},
  journal = {Nature Machine Intelligence},
  volume = {2},
  number = {11},
  pages = {665--673},
  year = {2020},
  doi = {10.1038/s42256-020-00257-z},
  url = {https://www.nature.com/articles/s42256-020-00257-z}
}

@inproceedings{matton2025walk,
  title = {Walk the Talk? Measuring the Faithfulness of Large Language Model Explanations},
  author = {Matton, Katie and Ness, Robert and Guttag, John and Kiciman, Emre},
  booktitle = {International Conference on Learning Representations},
  year = {2025},
  url = {https://proceedings.iclr.cc/paper_files/paper/2025/hash/b5ec50eb177908f21f78ed0d76ed525c-Abstract-Conference.html}
}

@inproceedings{jiang2025personamem,
  title = {Know Me, Respond to Me: Benchmarking {LLM}s for Dynamic User Profiling and Personalized Responses at Scale},
  author = {Jiang, Bowen and Hao, Zhuoqun and Cho, Young-Min and Li, Bryan and Yuan, Yuan and Chen, Sihao and Ungar, Lyle and Taylor, Camillo J. and Roth, Dan},
  booktitle = {Second Conference on Language Modeling},
  year = {2025},
  url = {https://openreview.net/forum?id=6ox8XZGOqP}
}

@inproceedings{shen2026mem2actbench,
  title = {{Mem2ActBench}: A Benchmark for Evaluating Long-Term Memory Utilization in Task-Oriented Autonomous Agents},
  author = {Shen, Yiting and Li, Kun and Zhou, Wei and Hu, Songlin},
  booktitle = {Proceedings of the 64th Annual Meeting of the Association for Computational Linguistics (Volume 1: Long Papers)},
  pages = {8173--8190},
  year = {2026},
  publisher = {Association for Computational Linguistics},
  doi = {10.18653/v1/2026.acl-long.370},
  url = {https://aclanthology.org/2026.acl-long.370/}
}

@inproceedings{meng2022memit,
  title = {Mass-Editing Memory in a Transformer},
  author = {Meng, Kevin and Sharma, Arnab Sen and Andonian, Alex and Belinkov, Yonatan and Bau, David},
  booktitle = {International Conference on Learning Representations},
  year = {2023},
  url = {https://openreview.net/forum?id=MkbcAHIYgyS}
}

@article{kadavath2022know,
  title = {Language Models (Mostly) Know What They Know},
  author = {Kadavath, Saurav and Conerly, Tom and Askell, Amanda and Henighan, Tom and Drain, Dawn and Perez, Ethan and Schiefer, Nicholas and Hatfield-Dodds, Zac and DasSarma, Nova and Tran-Johnson, Eli and others},
  journal = {arXiv preprint arXiv:2207.05221},
  year = {2022},
  url = {https://arxiv.org/abs/2207.05221}
}

\clearpage
\onecolumn
\appendix
\makeatletter
\renewenvironment{table}[1][]{\@float{table}[H]}{\end@float}
\renewenvironment{figure}[1][]{\@float{figure}[H]}{\end@float}
\makeatother
Appendix section headings retain artifact identifiers; Appendix~\ref{sec:app-artifact-name-mapping} maps them to reader-facing names.
\section{Controlled V3 Construction and Validation}
Frozen V3 contains 160 synthetically generated examples arranged as 40 four-way scenario families. Each family has one \textsc{Use}, one \textsc{Ignore}, one \textsc{Update}, and one \textsc{Ask} variant. The task string is byte-identical across the four variants, and each example has exactly four history records: an earlier temporal record, a later temporal record, and two context-specific records. Reference policies are derived from structured semantic relations before rendering, not assigned by inspecting model outputs.

The approved frozen set is rule-derived, deterministically validated, and was verified by two annotators, working separately and blind to the rule-derived reference policies, who agreed with each other and with the reference policy on all 160 examples. Annotation was not performed by an independent third party, and labels remain rule-derived. Structural validation checks policy balance, four records per example, identical within-family task strings, type-compatible slot values, template balance, duplicate and near-duplicate diagnostics, and absence of direct policy/state label leakage.

\section{Frozen-V3 Matched-Ablation Prompts}
\label{app:frozen-v3-matched-prompts}

All matched-ablation arms instantiate the same history/task prefix:
\begin{PromptBlock}
User History:
{history}
Current Task: {task}
\end{PromptBlock}

The policy definitions are:
\begin{PromptBlock}
- Use: apply the relevant memory.
- Ignore: complete the task without the memory.
- Update: acknowledge that newer information supersedes the old memory and use the newer information.
- Ask: request clarification before acting.
\end{PromptBlock}

When a state taxonomy is included, the matched-ablation runner uses this exact heading and definitions:
\begin{PromptBlock}
Memory state options:
- Active: a relevant memory is current and applies to this task.
- Stale_or_Irrelevant: the memory should not be applied and no clarification is needed.
- Conflicting_or_Superseded: newer information explicitly overrides older memory.
- Underspecified: the history does not contain enough information to decide.
\end{PromptBlock}

\paragraph{\textsc{Clean\_Policy\_Only}.}
\begin{PromptBlock}
User History:
{history}
Current Task: {task}

Policy options:
{policy definitions}

Output strictly valid JSON with no markdown:
{"policy": "Use|Ignore|Update|Ask"}
\end{PromptBlock}

\paragraph{\textsc{Taxonomy\_No\_State\_Output}.}
\begin{PromptBlock}
User History:
{history}
Current Task: {task}

Memory state options:
{state definitions}

Policy options:
{policy definitions}

Output strictly valid JSON with no markdown:
{"policy": "Use|Ignore|Update|Ask"}
\end{PromptBlock}

For auditability, the following is the verbatim taxonomy-only prompt rendered by the existing runner for example \texttt{controlled\_v3\_02\_use}.
\begin{PromptBlock}
User History:
Jordan saved an earlier entry: flight seat choice uses an aisle seat for its flight seat choice option.
The later note records an aisle seat for the flight seat choice option of flight seat choice.
Case A: Jordan noted that the train seat option for train seat is a window seat.
Case B: the profile lists I sit near an exit as the train seat option for train seat.
Current Task: Choose the preference option for Jordan's flight seat choice.

Memory state options:
- Active: a relevant memory is current and applies to this task.
- Stale_or_Irrelevant: the memory should not be applied and no clarification is needed.
- Conflicting_or_Superseded: newer information explicitly overrides older memory.
- Underspecified: the history does not contain enough information to decide.

Policy options:
- Use: apply the relevant memory.
- Ignore: complete the task without the memory.
- Update: acknowledge that newer information supersedes the old memory and use the newer information.
- Ask: request clarification before acting.

Output strictly valid JSON with no markdown:
{"policy": "Use|Ignore|Update|Ask"}
\end{PromptBlock}

\paragraph{\textsc{State\_Output\_No\_Rationale}.}
\begin{PromptBlock}
User History:
{history}
Current Task: {task}

Memory state options:
{state definitions}

Policy options:
{policy definitions}

Output strictly valid JSON with no markdown:
{"state": "Active|Stale_or_Irrelevant|Conflicting_or_Superseded|Underspecified",
 "policy": "Use|Ignore|Update|Ask"}
\end{PromptBlock}

\paragraph{\textsc{State\_Output\_With\_Rationale}.}
\begin{PromptBlock}
User History:
{history}
Current Task: {task}

Memory state options:
{state definitions}

Policy options:
{policy definitions}

Step 1. Classify the memory state.
Step 2. Choose the policy.
Output strictly valid JSON with no markdown:
{"state": "Active|Stale_or_Irrelevant|Conflicting_or_Superseded|Underspecified",
 "reasoning": "...",
 "policy": "Use|Ignore|Update|Ask"}
\end{PromptBlock}

\paragraph{\textsc{State\_Only\_Deterministic\_Routing}.}
\begin{PromptBlock}
User History:
{history}
Current Task: {task}

Memory state options:
{state definitions}

Policy options:
{policy definitions}

Output strictly valid JSON with no markdown:
{"state": "Active|Stale_or_Irrelevant|Conflicting_or_Superseded|Underspecified"}
\end{PromptBlock}

\section{Frozen-V3 State-Intervention Prompts}
\label{app:frozen-v3-label-conditioning-prompts}

The label-conditioning diagnostic prompts below are frozen-V3 prompts, not historical development-set internal prompts. They share the same policy definitions listed in the matched-ablation prompt section; conditions that ask for a state use the same state definitions.

\paragraph{\textsc{Policy\_Only}.}
\begin{PromptBlock}
User History:
{history}
Current Task: {task}

Policy options:
{policy definitions}

Output JSON: {"policy": "Use|Ignore|Update|Ask"}
\end{PromptBlock}

\paragraph{\textsc{State\_First}.}
\begin{PromptBlock}
User History:
{history}
Current Task: {task}

State options:
{state definitions}

Policy options:
{policy definitions}

First output state, then policy. Output JSON: {"state": "...", "policy": "..."}
\end{PromptBlock}

\paragraph{\textsc{Policy\_First}.}
\begin{PromptBlock}
User History:
{history}
Current Task: {task}

State options:
{state definitions}

Policy options:
{policy definitions}

First output policy, then state. Output JSON: {"policy": "...", "state": "..."}
\end{PromptBlock}

\paragraph{\textsc{Forced\_Correct\_State}.}
\begin{PromptBlock}
User History:
{history}
Current Task: {task}

Given memory state: {benchmark-implied state}

Policy options:
{policy definitions}

Output JSON: {"policy": "Use|Ignore|Update|Ask"}
\end{PromptBlock}

\paragraph{\textsc{Forced\_Wrong\_State}.}
\begin{PromptBlock}
User History:
{history}
Current Task: {task}

Given memory state: {fixed non-reference state}

Policy options:
{policy definitions}

Output JSON: {"policy": "Use|Ignore|Update|Ask"}
\end{PromptBlock}

\paragraph{\textsc{State\_Only\_Deterministic\_Routing}.}
\begin{PromptBlock}
User History:
{history}
Current Task: {task}

State options:
{state definitions}

Output JSON:
{"state": "Active|Stale_or_Irrelevant|Conflicting_or_Superseded|Underspecified"}
\end{PromptBlock}

\section{Artifact Name Mapping}\label{sec:app-artifact-name-mapping}
Table~\ref{tab:app-artifact-name-mapping} reports the bridge between reader-facing names and artifact identifiers.
\begin{table}[H]
\centering
\small
\begin{tabular}{>{\raggedright\arraybackslash}p{0.36\columnwidth}>{\raggedright\arraybackslash}p{0.54\columnwidth}}
\toprule
Paper-facing name & Artifact ID \\
\midrule
Policy-only matched arm & Clean\_Policy\_Only \\
Explicit state-output matched arm & State\_Output\_No\_Rationale \\
State-only routing with policy definitions & State\_Only\_Deterministic\_Routing (matched-ablation prompt) \\
State-only routing without policy definitions & State\_Only\_Deterministic\_Routing (label-conditioning prompt) \\
Supplied reference label & Forced\_Correct\_State \\
Supplied conflicting label & Forced\_Wrong\_State \\
the frozen controlled set & Frozen V3 \\
the semantic-evidence follow-up & Semantic-Evidence V1 \\
\bottomrule
\end{tabular}
\caption{Reader-facing names for internal artifact identifiers. Exact machine IDs are stored in the raw-result metadata. The two state-only deterministic-routing artifact conditions share a historical ID but use different prompts; they are therefore named separately in prose.}
\label{tab:app-artifact-name-mapping}
\end{table}

\section{Label-Conditioning Diagnostic and Wrong-State Construction}\label{sec:app-label-conditioning-wrong-state}
The \textsc{Forced\_Correct\_State} and \textsc{Forced\_Wrong\_State} artifact conditions are reported as a label-conditioning sensitivity diagnostic. Frozen V3 uses a deterministic benchmark state-to-policy mapping: Active$\rightarrow$\textsc{Use}, Stale-or-Irrelevant$\rightarrow$\textsc{Ignore}, Conflicting-or-Superseded$\rightarrow$\textsc{Update}, and Underspecified$\rightarrow$\textsc{Ask}. Supplying a state label therefore supplies a benchmark-associated answer label, not an independent latent mechanism.

The \textsc{Forced\_Wrong\_State} condition uses the fixed seed 20270719. For each frozen example, the runner samples one label uniformly from the three states that differ from the benchmark-implied state. The selected wrong state is then fixed across models and seeds. This construction makes wrong-state comparisons informative about sensitivity to externally supplied benchmark-associated labels, but it does not prove that naturally emitted state strings are faithful internal mechanisms.

\section{Full Frozen-V3 Results}
Table~\ref{tab:app-frozen-v3-matched-full} reports the full matched prompt-ablation results, and Table~\ref{tab:app-frozen-v3-state-full} reports the full label-conditioning diagnostic results.
\begin{table}[H]
\centering
\small
\resizebox{\linewidth}{!}{%
\begin{tabular}{llrrrr}
\toprule
Model & Arm & Calls & Accuracy & Macro F1 & Parse fail \\
\midrule
Llama-3.3-70B & \textsc{CLEAN\_POLICY\_ONLY} & 480 & 44.0\% & 36.3\% & 0 \\
Llama-3.3-70B & \textsc{TAXONOMY\_NO\_STATE\_OUTPUT} & 480 & 53.1\% & 46.5\% & 1 \\
Llama-3.3-70B & \textsc{STATE\_OUTPUT\_NO\_RATIONALE} & 480 & 44.6\% & 35.4\% & 0 \\
Llama-3.3-70B & \textsc{STATE\_OUTPUT\_WITH\_RATIONALE} & 480 & 47.9\% & 40.4\% & 0 \\
Llama-3.3-70B & \textsc{STATE\_ONLY\_DETERMINISTIC\_ROUTING} & 480 & 45.0\% & 35.9\% & 0 \\
GPT-OSS-120B & \textsc{CLEAN\_POLICY\_ONLY} & 480 & 54.8\% & 47.1\% & 0 \\
GPT-OSS-120B & \textsc{TAXONOMY\_NO\_STATE\_OUTPUT} & 480 & 59.8\% & 55.2\% & 0 \\
GPT-OSS-120B & \textsc{STATE\_OUTPUT\_NO\_RATIONALE} & 480 & 58.1\% & 53.5\% & 0 \\
GPT-OSS-120B & \textsc{STATE\_OUTPUT\_WITH\_RATIONALE} & 480 & 57.9\% & 52.9\% & 0 \\
GPT-OSS-120B & \textsc{STATE\_ONLY\_DETERMINISTIC\_ROUTING} & 480 & 56.9\% & 50.8\% & 0 \\
\bottomrule
\end{tabular}%
}
\caption{Full frozen-V3 matched prompt-ablation results.}
\label{tab:app-frozen-v3-matched-full}
\end{table}

\begin{table}[H]
\centering
\small
\resizebox{\textwidth}{!}{%
\begin{tabular}{llrrrrr}
\toprule
Model & Condition & Calls & Accuracy & Macro F1 & State agreement & Parse fail \\
\midrule
Llama-3.3-70B & \textsc{POLICY\_ONLY} & 480 & 45.4\% & 36.3\% & n/a & 0 \\
Llama-3.3-70B & \textsc{STATE\_FIRST} & 480 & 42.7\% & 32.6\% & 41.5\% & 0 \\
Llama-3.3-70B & \textsc{POLICY\_FIRST} & 480 & 48.8\% & 40.1\% & 44.6\% & 0 \\
Llama-3.3-70B & \textsc{FORCED\_CORRECT\_STATE} & 480 & 52.7\% & 49.2\% & n/a & 0 \\
Llama-3.3-70B & \textsc{FORCED\_WRONG\_STATE} & 480 & 37.5\% & 30.4\% & n/a & 0 \\
Llama-3.3-70B & \textsc{STATE\_ONLY\_DETERMINISTIC\_ROUTING} & 480 & 46.9\% & 36.1\% & 46.9\% & 0 \\
GPT-OSS-120B & \textsc{POLICY\_ONLY} & 480 & 53.1\% & 44.6\% & n/a & 0 \\
GPT-OSS-120B & \textsc{STATE\_FIRST} & 480 & 59.2\% & 54.4\% & 55.8\% & 0 \\
GPT-OSS-120B & \textsc{POLICY\_FIRST} & 480 & 57.1\% & 51.1\% & 53.8\% & 0 \\
GPT-OSS-120B & \textsc{FORCED\_CORRECT\_STATE} & 480 & 65.0\% & 63.2\% & n/a & 0 \\
GPT-OSS-120B & \textsc{FORCED\_WRONG\_STATE} & 480 & 50.0\% & 46.1\% & n/a & 0 \\
GPT-OSS-120B & \textsc{STATE\_ONLY\_DETERMINISTIC\_ROUTING} & 480 & 52.5\% & 46.5\% & 52.5\% & 0 \\
\bottomrule
\end{tabular}%
}
\caption{Full frozen-V3 label-conditioning diagnostic results. State agreement is shown only for conditions that ask the model to emit a state. Forced-state conditions supply benchmark-associated labels and are not interpreted as internal state mechanisms.}
\label{tab:app-frozen-v3-state-full}
\end{table}

\subsection{Per-Policy Decomposition of the Matched Ablation}
Table~\ref{tab:app-per-policy-decomposition} decomposes the primary matched ablation by reference policy.

\begin{table}[H]
\centering
\scriptsize
\setlength{\tabcolsep}{1pt}
\begin{tabular}{@{}>{\raggedright\arraybackslash}p{0.21\columnwidth}>{\raggedright\arraybackslash}p{0.31\columnwidth}>{\raggedright\arraybackslash}p{0.13\columnwidth}>{\raggedright\arraybackslash}p{0.18\columnwidth}>{\raggedright\arraybackslash}p{0.12\columnwidth}@{}}
\toprule
Model & Policy & $\Delta$ & 95\% CI & Holm $p$ \\
\midrule
Llama-3.3-70B & \textsc{Use} & +17.5 pp$^{*}$\, & [6.67, 28.33] & 0.010 \\
Llama-3.3-70B & \textsc{Ignore} & n.s. & n.s. & n.s. \\
Llama-3.3-70B & \textsc{Update} & n.s. & n.s. & n.s. \\
Llama-3.3-70B & \textsc{Ask} & -14.2 pp$^{*}$\, & [-22.50, -6.67] & 0.010 \\
GPT-OSS-120B & \textsc{Use} & n.s. & n.s. & n.s. \\
GPT-OSS-120B & \textsc{Ignore} & +15.8 pp$^{*}$\, & [7.50, 25.00] & 0.003 \\
GPT-OSS-120B & \textsc{Update} & -8.3 pp$^{*}$\, & [-13.33, -3.33] & 0.010 \\
GPT-OSS-120B & \textsc{Ask} & n.s. & n.s. & n.s. \\
\bottomrule
\end{tabular}
\caption{Per-policy decomposition of the primary matched ablation. Deltas are \textsc{State\_Output\_No\_Rationale} minus \textsc{Clean\_Policy\_Only}; CIs are scenario-family-cluster bootstrap intervals; $p$ is Holm-adjusted across the four policies within each model; parse failures count as incorrect. Only Holm-significant cells are numerically reported here; full per-policy metrics and confusion matrices appear below. Gains and losses are reported together and the overall effect is null. $^{*}$ marks Holm-significant cells.}
\label{tab:app-per-policy-decomposition}
\end{table}

\subsection{Third-Endpoint Replication (llama-3.1-8b-instant)}\label{sec:app-third-endpoint-replication}
Table~\ref{tab:app-third-endpoint-replication} reports an additive third-endpoint replication.

\begin{table}[H]
\centering
\scriptsize
\setlength{\tabcolsep}{1pt}
\begin{tabular}{@{}>{\raggedright\arraybackslash}p{0.21\columnwidth}>{\raggedright\arraybackslash}p{0.31\columnwidth}>{\raggedright\arraybackslash}p{0.13\columnwidth}>{\raggedright\arraybackslash}p{0.18\columnwidth}>{\raggedright\arraybackslash}p{0.12\columnwidth}@{}}
\toprule
Model & Arm & Accuracy & Macro F1 & Parse fail \\
\midrule
llama-3.1-8b-instant & Clean policy-only & 25.6\% & 11.3\% & 0/480 \\
llama-3.1-8b-instant & State-output no rationale & 37.3\% & 27.4\% & 18/480 \\
llama-3.1-8b-instant & Delta (state-output minus policy-only) & +11.7 pp & 95\% CI [8.54, 14.79] & p=0.0001 \\
\bottomrule
\end{tabular}
\caption{Additive third-endpoint replication. The candidate qwen/qwen3.6-27b was excluded after an HTTP 400 json\_validate\_failed smoke failure, mirroring the GPT-OSS explicit-evidence provider failure documented in the main text. llama-3.1-8b-instant is a small model included only to test cross-model direction, not as a headline result. The state-output arm had 18/480 parse failures, counted as incorrect, so this delta must not be read as a clean gain.}
\label{tab:app-third-endpoint-replication}
\end{table}

\section{Family-Level Counterfactual Analysis}
Frozen V3 is arranged as 40 matched four-way families. Family exact match
requires all four variants in a family to be correct for a given model, arm,
and seed. The analysis also constructs the six within-family policy pairs
(\textsc{Use}/\textsc{Ignore}, \textsc{Use}/\textsc{Update},
\textsc{Use}/\textsc{Ask}, \textsc{Ignore}/\textsc{Update},
\textsc{Ignore}/\textsc{Ask}, and \textsc{Update}/\textsc{Ask}) and classifies
each pair as expected-transition correct, partial transition, missed
transition, wrong-direction transition, or provider/parse failure. The computed
pairwise-transition records are retained in local run outputs.
Table~\ref{tab:app-family-counterfactual} reports family exact-match rates for the matched prompt ablation.
Table~\ref{tab:app-family-counterfactual-diagnostic} reports family exact-match rates for the label-conditioning diagnostic.
\begin{table}[H]
\centering
\small
\setlength{\tabcolsep}{4pt}
\begin{tabular}{llc}
\toprule
Model & Condition & Family exact \\
\midrule
Llama & Matched policy-only & 0.0\% [0.0, 2.5] \\
Llama & State output & 0.0\% [0.0, 2.5] \\
GPT-OSS & Matched policy-only & 1.7\% [0.0, 5.0] \\
GPT-OSS & State output & 5.0\% [0.8, 10.0] \\
\bottomrule
\end{tabular}
\caption{Family exact-match rates on the frozen controlled set. A family is correct only when all four matched variants are predicted correctly. Intervals resample scenario families, with seeds averaged inside each family. For 0.0\% entries, the 3/n upper bound uses n=120 family-seed units, an optimistic independence assumption.}
\label{tab:app-family-counterfactual}
\end{table}

\begin{table}[H]
\centering
\small
\setlength{\tabcolsep}{4pt}
\begin{tabular}{llc}
\toprule
Model & Condition & Family exact \\
\midrule
Llama & Diagnostic policy-only & 0.0\% [0.0, 2.5] \\
Llama & Supplied reference label & 1.7\% [0.0, 4.2] \\
Llama & Supplied conflicting label & 0.0\% [0.0, 2.5] \\
GPT-OSS & Diagnostic policy-only & 1.7\% [0.0, 4.2] \\
GPT-OSS & Supplied reference label & 11.7\% [5.0, 19.2] \\
GPT-OSS & Supplied conflicting label & 0.8\% [0.0, 2.5] \\
\bottomrule
\end{tabular}
\caption{Family exact-match rates for the separately executed label-conditioning diagnostic. The policy-only row comes from a different frozen protocol source and is labeled by protocol to avoid conflation. Intervals resample scenario families, with seeds averaged inside each family. For 0.0\% entries, the 3/n upper bound uses n=120 family-seed units, an optimistic independence assumption.}
\label{tab:app-family-counterfactual-diagnostic}
\end{table}

Pairwise transition analysis gives a second view. Across the six within-family policy pairs, expected-transition accuracy is 13.8\% for Llama policy-only and 13.6\% for Llama state-output; missed-transition rates are 48.6\% and 47.1\%. GPT-OSS improves from 23.5\% to 29.2\% expected transitions with state output, but still misses 32.8\% and 30.3\% of transitions. Supplied reference labels improve expected transitions relative to policy-only, again consistent with answer-associated label sensitivity rather than an internally faithful state mechanism.

\section{Seed-Stability Diagnostics}
Each hosted arm uses seeds 101, 102, and 103. We report whether the three
policies for an example are unanimous, a 2--1 majority, or all different, and
compute a simple entropy over the three predictions. Because only three seeds
are available, this is a stochastic-instability diagnostic rather than a
calibrated uncertainty estimate. The corresponding per-example outputs are
retained in local run outputs.

Among unanimous examples, per-call accuracy is 50.5\% and 48.4\% for Llama policy-only and state-output, and 61.7\% and 63.9\% for GPT-OSS.
Non-unanimous examples often have lower per-call accuracy, but the pattern is descriptive and not calibrated uncertainty.

\section{Normalized-Input Probe Protocol}\label{sec:app-normalized-input-protocol}

The normalized-input probe uses the same final controlled set as the matched experiments; the controlled set records 160 examples and 40 families. The normalized-input policy-only arm is compared against clean policy-only. It uses seeds 101, 102, and 103, temperature 0.7, and the two endpoints \texttt{llama-3.3-70b-versatile} and \texttt{openai/gpt-oss-120b}. The normalized-input arm records 480 calls and 0 parse failures for each normalized-input model-arm, 960 calls total.

The tuple block is derived deterministically from the stored semantic records: for each record, the renderer writes subject, attribute, value, and either the stored time tag, stored context tag, or \texttt{none}. The arm keeps the clean policy-only task and replaces the free-text history with deterministic structured tuple lines in the \texttt{[subject|attribute|value|tag]} format. The rendered prompt below was produced by the existing runner for example \texttt{controlled\_v3\_02\_use}.

\begin{PromptBlock}
User History:
[flight seat choice | flight seat choice option | an aisle seat | earlier]
[flight seat choice | flight seat choice option | an aisle seat | later]
[train seat | train seat option | a window seat | case A]
[train seat | train seat option | I sit near an exit | case B]
Current Task: Choose the preference option for Jordan's flight seat choice.

Policy options:
- Use: apply the relevant memory.
- Ignore: complete the task without the memory.
- Update: acknowledge that newer information supersedes the old memory and use the newer information.
- Ask: request clarification before acting.

Output strictly valid JSON with no markdown:
{"policy": "Use|Ignore|Update|Ask"}
\end{PromptBlock}

The leakage guard trains grouped word and character TF-IDF decoders on the normalized tuple block plus task text, grouping by scenario family. The leakage probe reports word-grouped accuracy 0.375, character-grouped accuracy 0.33125, a near-trivial decode threshold of 0.9, and was not flagged as answer leaking.

\begin{table}[H]
\centering
\small
\resizebox{\linewidth}{!}{%
\begin{tabular}{lrrrrrr}
\toprule
Model & Accuracy & Macro F1 & Parse failures & Delta vs clean & 95\% CI & $p$ \\
\midrule
\texttt{llama-3.3-70b-versatile} & 35.42\% & 30.36\% & 0 & -8.54 pp & [-13.96, -3.54] & 0.003400 \\
\texttt{openai/gpt-oss-120b} & 46.88\% & 44.08\% & 0 & -7.92 pp & [-12.08, -3.54] & 0.001200 \\
\bottomrule
\end{tabular}}
\caption{Normalized-input probe results. Deltas are normalized-input minus clean policy-only.}
\label{tab:app-normalized-input-results}
\end{table}

The reported normalized-input values are paired p-values; no Holm-adjusted values for this probe are reported here. Across the paper, Holm-adjusted values are reported where multiple contrasts are grouped as a family of tests, and otherwise the reported p-values are treated as descriptive exploratory evidence.

\section{Rule-Based Error Taxonomy}\label{sec:app-error-taxonomy}
The error taxonomy is automatically derived from structured metadata and model
predictions. It is not a human-coded annotation. Categories include recency
overreach, conflict over-detection, irrelevance-to-\textsc{Ask} confusion,
ambiguity under-detection, defaulting to \textsc{Use} or \textsc{Ask},
state-policy inconsistency, intermediate serialization burden, task neglect,
and provider/protocol failure. Categories are non-exclusive, may overlap, and
are counted over final scored model-arm-example-seed records rather than raw
retry attempts. The local error-analysis outputs retain full examples,
denominators, and rates.
Table~\ref{tab:app-error-taxonomy} reports the rule-based error-mode counts.
\begin{table}[H]
\centering
\small
\setlength{\tabcolsep}{4pt}
\begin{tabular}{@{}>{\raggedright\arraybackslash}p{0.45\columnwidth}rr>{\raggedright\arraybackslash}p{0.18\columnwidth}@{}}
\toprule
Error mode & Count & Eligible & Detail \\
\midrule
conflict over detection & 2428 & 5121 &  \\
task neglect & 1583 & 5121 &  \\
recency overreach & 1543 & 5121 &  \\
ask defaulting & 873 & 5121 &  \\
irrelevance to ask confusion & 760 & 5121 &  \\
state policy inconsistency & 276 & 3840 & 138 final policy correct \\
use defaulting & 266 & 5121 &  \\
ambiguity under detection & 123 & 5121 &  \\
\bottomrule
\end{tabular}
\caption{Rule-based error-mode counts across final scored controlled-set model-arm-example-seed records. Eligible denotes the count of final scored records for which the category is applicable (5121 for arms scored on all records; 3840 for state-emitting arms only). Counts are non-exclusive and must not be summed.}
\label{tab:app-error-taxonomy}
\end{table}

\section{Per-Policy Metrics and Confusion Matrices}
The machine-readable frozen-V3 reports include per-policy precision, recall, F1, and confusion matrices for each model and arm. Parse failures remain explicit records and count as incorrect. The main paper reports only the matched headline contrasts to preserve readability.
Table~\ref{tab:app-per-policy-prf} reports per-policy precision, recall, and F1 for the two primary matched-ablation arms.

\begin{table}[H]
\centering
\small
\begin{tabular}{lllrrr}
\toprule
Model & Arm & Policy & Precision & Recall & F1 \\
\midrule
Llama-3.3-70B & Clean policy-only & \textsc{Use} & 85.1 & 61.7 & 71.5 \\
Llama-3.3-70B & Clean policy-only & \textsc{Ignore} & 100.0 & 0.8 & 1.7 \\
Llama-3.3-70B & Clean policy-only & \textsc{Update} & 34.6 & 98.3 & 51.2 \\
Llama-3.3-70B & Clean policy-only & \textsc{Ask} & 35.3 & 15.0 & 21.1 \\
Llama-3.3-70B & State-output & \textsc{Use} & 84.8 & 79.2 & 81.9 \\
Llama-3.3-70B & State-output & \textsc{Ignore} & 83.3 & 4.2 & 7.9 \\
Llama-3.3-70B & State-output & \textsc{Update} & 34.6 & 94.2 & 50.6 \\
Llama-3.3-70B & State-output & \textsc{Ask} & 2.9 & 0.8 & 1.3 \\
GPT-OSS-120B & Clean policy-only & \textsc{Use} & 98.3 & 99.2 & 98.8 \\
GPT-OSS-120B & Clean policy-only & \textsc{Ignore} & 100.0 & 9.2 & 16.8 \\
GPT-OSS-120B & Clean policy-only & \textsc{Update} & 40.7 & 100.0 & 57.8 \\
GPT-OSS-120B & Clean policy-only & \textsc{Ask} & 24.5 & 10.8 & 15.0 \\
GPT-OSS-120B & State-output & \textsc{Use} & 90.2 & 100.0 & 94.9 \\
GPT-OSS-120B & State-output & \textsc{Ignore} & 100.0 & 25.0 & 40.0 \\
GPT-OSS-120B & State-output & \textsc{Update} & 40.1 & 91.7 & 55.8 \\
GPT-OSS-120B & State-output & \textsc{Ask} & 44.2 & 15.8 & 23.3 \\
\bottomrule
\end{tabular}
\caption{Per-policy precision, recall, and F1 for the two primary matched-ablation arms. Values are percentages. Parse failures count as incorrect.}
\label{tab:app-per-policy-prf}
\end{table}

\section{Statistical Procedures}
The unit of clustering is the scenario family. Reported confidence intervals use a 10,000-resample percentile bootstrap over scenario-family clusters. Paired tests use family-level sign-flip permutation tests with plus-one correction. Non-significant p-values are not interpreted as equivalence. The primary matched-ablation contrast is state-output minus clean policy-only. In the label-conditioning diagnostic, forced-correct-state, forced-wrong-state, and state-first conditions are compared against policy-only, and wrong-state routing is also compared against forced-correct routing.

\section{Shortcut-Transfer Diagnostic}\label{sec:app-shortcut-transfer}
A supervised TF-IDF classifier trained only on the original 480-example synthetic development set reaches 25.0\% accuracy and 13.9\% macro F1 when evaluated once on frozen V3. The classifier vocabulary and parameters are fit only on the development set. Because the development set has no positive \textsc{Ignore} examples, this transfer result is affected by label-space shift, class-prior shift, and template shift. It is a descriptive transfer diagnostic, not the primary evidence that frozen V3 eliminates lexical shortcuts.

\section{Semantic-Evidence V1 Follow-Up}\label{sec:app-semantic-evidence-followup}
The semantic-evidence v1 protocol was created after identifying the deterministic mapping between frozen-V3 benchmark states and policies. It uses 160 examples arranged as 40 matched families and three prompt arms: policy-only, explicit semantic-evidence output, and evidence-taxonomy exposure without evidence output. The explicit-evidence arm asks for task relevance, temporal status, supersession, context sufficiency, risk or confirmation, and policy. It does not supply the old four-state benchmark label to the model.

The local design audit passes structural checks. Single evidence-field majority decode accuracies range from 25.0\% to 50.0\%; the full evidence tuple decodes policy at 75.0\%, template signatures at 75.0\%, keyword counts at 51.2\%, grouped word TF-IDF at 55.6\% accuracy and 55.6\% macro F1, and grouped character TF-IDF at 62.5\% accuracy and 61.7\% macro F1. These diagnostics are reported as design constraints, not as model-performance results.

All 2,880 hosted calls in the frozen manifest were attempted. Llama completed all calls without provider failures. Its policy-only, taxonomy-only, and explicit semantic-evidence accuracies are 63.3\%, 54.6\%, and 55.8\%, respectively. Explicit semantic evidence is 7.5 points below policy-only (95\% CI [-12.7, -2.5], Holm-adjusted p=0.0146), 1.3 points above taxonomy-only (p=0.6079), and taxonomy-only is 8.8 points below policy-only (95\% CI [-12.1, -5.6], Holm-adjusted p=0.0003). GPT-OSS produced provider-side JSON validation failures for all explicit-evidence calls, so that arm is not a clean behavioral comparison for GPT-OSS. The failed records are retained in execution accounting; no prompt, schema, or parsing rule was changed after observing them, and the GPT-OSS explicit-evidence arm is not scored as a clean model-behavior result.

\section{Proposal-Only Future Protocols}
Two optional future protocols were drafted but not run: a downstream-consequence
proposal and an independent-renderer proposal. They contain no reported result
numbers, require separate explicit API authorization before any hosted
execution, and are not used to support current manuscript claims.

\section{Model-Protocol Amendment}
The development-set evidence used Llama-3.3-70B and Qwen3-32B. Before any frozen-V3 hosted result existed, the planned Qwen3-32B endpoint became unavailable on the current Groq tier. The frozen-V3 protocol was therefore amended to use Llama-3.3-70B and GPT-OSS-120B. Frozen data, prompts, arms, seeds, scoring, and analysis rules were not changed, and GPT-OSS results are never labeled as Qwen results.

\section{Historical Development-Set Results}\label{sec:app-historical-development-results}
The original 480-example development set remains useful as diagnostic evidence but not as the decisive test. It has positive reference-policy examples for \textsc{Use}, \textsc{Update}, and \textsc{Ask}, and no standalone positive \textsc{Ignore} reference class. It is also lexically separable under supervised TF-IDF diagnostics. Historical prompt-bundle results and development-set confusion matrices are retained below for auditability. \textsc{Full-context} uses the full user history if relevant and then asks the model to decide the best policy. \textsc{Ask-if-uncertain} instructs the model to choose \textsc{Ask} when uncertain whether the user history applies; otherwise it asks the model to decide the best policy.

% Intentionally empty. Supplemental tables are input directly from generated
% artifact files to avoid an otherwise empty appendix divider page.

Table~\ref{tab:app-state_confusion} reports the state confusion matrix for the state-structured bundle.
\begin{table}[H]
\centering
\footnotesize
\caption{\textsc{State-structured bundle} state confusion matrix pooled over three seeds. Rows are gold benchmark categories; columns are predicted memory states. Overall benchmark-mapped state agreement under the mapping Stable/Contextual$\rightarrow$Active, Updated$\rightarrow$Conflicting, and Ambiguous$\rightarrow$Underspecified is 79.4\% for Llama-3.3-70B and 73.1\% for Qwen3-32B.}
\label{tab:app-state_confusion}
\resizebox{\columnwidth}{!}{%
\begin{tabular}{llrrrrrr}
\toprule
Model & Generation category & Active & Stale & Conflicting & Underspecified & Parse fail & $n$ \\
\midrule
Llama-3.3-70B & Stable & 348 & 0 & 3 & 9 & 0 & 360 \\
Llama-3.3-70B & Contextual & 343 & 0 & 5 & 12 & 0 & 360 \\
Llama-3.3-70B & Updated & 33 & 4 & 323 & 0 & 0 & 360 \\
Llama-3.3-70B & Ambiguous & 108 & 20 & 102 & 130 & 0 & 360 \\
\addlinespace[0.35em]
\midrule
Qwen3-32B & Stable & 354 & 0 & 1 & 3 & 2 & 360 \\
Qwen3-32B & Contextual & 358 & 0 & 0 & 2 & 0 & 360 \\
Qwen3-32B & Updated & 49 & 30 & 275 & 5 & 1 & 360 \\
Qwen3-32B & Ambiguous & 221 & 13 & 59 & 65 & 2 & 360 \\
\bottomrule
\end{tabular}
}
\end{table}

Table~\ref{tab:app-three_arm_ablation} reports the historical three-arm development-set ablation.
\begin{table}[H]
\centering
\small
\caption{Three-arm ablation comparing the historical implicit-state/no-state-output prompt, the cleaned implicit-state/no-state-output prompt, and the state-structured prompt bundle. $p$-values are pooled exact McNemar diagnostics over 1440 matched example-seed decisions, with parse failures counted as wrong; cluster-aware comparisons are reported separately.}
\label{tab:app-three_arm_ablation}
\resizebox{\linewidth}{!}{%
\begin{tabular}{llrrrrr}
\toprule
Model & Arm & Overall & Updated & Parse fail & $p$ vs Historical & $p$ vs Cleaned \\
\midrule
Llama-3.3-70B & \textsc{Historical no-state-output} & 70.3±0.5\% & 46.7±3.8\% & 0.0±0.0\% & -- & -- \\
Llama-3.3-70B & \textsc{Cleaned no-state-output} & 71.7±0.6\% & 48.6±3.2\% & 0.0±0.0\% & -- & -- \\
Llama-3.3-70B & \textsc{State-structured bundle} & 85.1±0.2\% & 89.7±2.7\% & 0.0±0.0\% & 3.6e-37 & 2.4e-32 \\
\midrule
Qwen3-32B & \textsc{Historical no-state-output} & 65.3±1.8\% & 30.3±5.4\% & 0.2±0.2\% & -- & -- \\
Qwen3-32B & \textsc{Cleaned no-state-output} & 64.3±0.8\% & 33.6±2.7\% & 0.0±0.0\% & -- & -- \\
Qwen3-32B & \textsc{State-structured bundle} & 75.5±1.2\% & 75.3±3.8\% & 0.3±0.3\% & 1.5e-18 & 2.9e-26 \\
\bottomrule
\end{tabular}%
}
\end{table}

Table~\ref{tab:app-state_oracle} reports the historical oracle-worded routing diagnostic.
\begin{table}[H]
\centering
\small
\caption{\textsc{State-structured bundle} versus \textsc{Historical oracle-worded routing}. This historical language-model routing diagnostic uses oracle-worded prompt text and should not be interpreted as a neutral mapped-state intervention or a deterministic oracle.}
\label{tab:app-state_oracle}
\resizebox{\textwidth}{!}{%
\begin{tabular}{llrrrrrr}
\toprule
Model & Arm & Overall & Stable & Contextual & Updated & Ambiguous & Parse fail \\
\midrule
Llama-3.3-70B & \textsc{State-structured bundle} & 85.1±0.2\% & 96.4±0.5\% & 95.8±0.8\% & 89.7±2.7\% & 58.3±2.2\% & 0.0±0.0\% \\
Llama-3.3-70B & \textsc{Historical oracle-worded routing} & 86.9±2.2\% & 99.2±0.0\% & 97.5±0.0\% & 72.5±3.0\% & 78.3±6.0\% & 0.0\% \\
\midrule
Qwen3-32B & \textsc{State-structured bundle} & 75.5±1.2\% & 98.6±1.3\% & 99.2±0.8\% & 75.3±3.8\% & 28.9±1.0\% & 0.3±0.3\% \\
Qwen3-32B & \textsc{Historical oracle-worded routing} & 73.5±2.1\% & 99.7±0.5\% & 100.0±0.0\% & 50.0±3.0\% & 44.2±6.6\% & 0.0\% \\
\bottomrule
\end{tabular}
}
\end{table}

Table~\ref{tab:app-cost_table} reports token cost and strict accuracy for direct policy prediction versus explicit state classification.
\begin{table}[H]
\centering
\scriptsize
\setlength{\tabcolsep}{1pt}
\caption{Token cost and strict accuracy for direct policy prediction versus explicit state classification. Token counts are aggregated over 1440 calls.}
\label{tab:app-cost_table}
\begin{tabular}{@{}>{\raggedright\arraybackslash}p{0.24\columnwidth}>{\raggedright\arraybackslash}p{0.32\columnwidth}rrr@{}}
\toprule
Model & Arm & Tokens & Tokens/call & Accuracy \\
\midrule
Llama-3.3-70B & \textsc{Direct policy (no classification)} & 219,903 & 152.7 & 69.8±1.6\% \\
Llama-3.3-70B & \textsc{State-structured bundle} & 475,302 & 330.1 & 85.1±0.2\% \\
\midrule
Qwen3-32B & \textsc{Direct policy (no classification)} & 563,470 & 391.3 & 66.7±1.7\% \\
Qwen3-32B & \textsc{State-structured bundle} & 811,965 & 563.9 & 75.5±1.2\% \\
\bottomrule
\end{tabular}
\end{table}

The 480-example development set remains useful for diagnosing why the stronger controlled evaluation was needed. On that set, the old state-structured prompt bundle improved exact-match policy accuracy over the historical implicit-state/no-state-output prompt by 14.8 points for Llama and 10.2 points for Qwen. However, the same dataset is lexically separable, lacks positive \textsc{Ignore} coverage, and used a prompt bundle that changed taxonomy, step wording, state output, rationale output, and length simultaneously. These facts demote the development result from a method claim to a motivation for controlled counterfactual testing.

\textbf{Why the controlled result differs from the development result.} The development-set state-structured bundle arm was a prompt bundle, not an isolated state-field test. It changed policy wording, taxonomy exposure, output order, rationale generation, and token budget. It was also evaluated on a lexically separable set without positive \textsc{Ignore} examples. The frozen controlled set balances all four policies and is evaluated with grouped diagnostics, so it is the stronger evidence for the state-field question; however, the development and frozen-set shortcut numbers are not a direct same-protocol magnitude comparison.

\section{Policy Confusion Matrices}
\label{app:confusion_matrices}

Table~\ref{tab:app-policy-confusions-1} and Table~\ref{tab:app-policy-confusions-2} report the development-set policy confusion matrices by endpoint.
Rows are generation categories and columns are predicted policies. All matrices use the full 480-example Groq protocol with seeds 101/102/103. The \textsc{Historical oracle-worded routing} rows use the historical oracle-worded routing prompt.

\begin{table}[H]
\centering
\small
\caption{Policy confusion matrix for Llama-3.3-70B. Rows are gold benchmark categories; columns are predicted policies plus parse failures.}
\label{tab:app-policy-confusions-1}
\resizebox{\linewidth}{!}{%
\begin{tabular}{llrrrrrr}
\toprule
Agent & Generation category & Use & Ask & Ignore & Update & Parse fail & $n$ \\
\midrule
\textsc{Historical no-state-output} & Stable & 355 & 2 & 1 & 2 & 0 & 360 \\
\textsc{Historical no-state-output} & Contextual & 356 & 2 & 2 & 0 & 0 & 360 \\
\textsc{Historical no-state-output} & Updated & 157 & 7 & 28 & 168 & 0 & 360 \\
\textsc{Historical no-state-output} & Ambiguous & 165 & 133 & 62 & 0 & 0 & 360 \\
\addlinespace[0.25em]
\textsc{Cleaned no-state-output} & Stable & 354 & 3 & 3 & 0 & 0 & 360 \\
\textsc{Cleaned no-state-output} & Contextual & 357 & 0 & 3 & 0 & 0 & 360 \\
\textsc{Cleaned no-state-output} & Updated & 165 & 6 & 14 & 175 & 0 & 360 \\
\textsc{Cleaned no-state-output} & Ambiguous & 146 & 146 & 67 & 1 & 0 & 360 \\
\addlinespace[0.25em]
\textsc{Direct policy (no classification)} & Stable & 336 & 12 & 0 & 12 & 0 & 360 \\
\textsc{Direct policy (no classification)} & Contextual & 355 & 1 & 3 & 1 & 0 & 360 \\
\textsc{Direct policy (no classification)} & Updated & 124 & 19 & 3 & 214 & 0 & 360 \\
\textsc{Direct policy (no classification)} & Ambiguous & 157 & 100 & 42 & 61 & 0 & 360 \\
\addlinespace[0.25em]
\textsc{State-structured bundle} & Stable & 347 & 11 & 0 & 2 & 0 & 360 \\
\textsc{State-structured bundle} & Contextual & 345 & 11 & 2 & 2 & 0 & 360 \\
\textsc{State-structured bundle} & Updated & 34 & 0 & 3 & 323 & 0 & 360 \\
\textsc{State-structured bundle} & Ambiguous & 109 & 210 & 34 & 7 & 0 & 360 \\
\addlinespace[0.25em]
\textsc{Historical oracle-worded routing} & Stable & 357 & 3 & 0 & 0 & 0 & 360 \\
\textsc{Historical oracle-worded routing} & Contextual & 351 & 0 & 9 & 0 & 0 & 360 \\
\textsc{Historical oracle-worded routing} & Updated & 34 & 56 & 9 & 261 & 0 & 360 \\
\textsc{Historical oracle-worded routing} & Ambiguous & 59 & 282 & 19 & 0 & 0 & 360 \\
\addlinespace[0.25em]
Ask-if-uncertain & Stable & 350 & 10 & 0 & 0 & 0 & 360 \\
Ask-if-uncertain & Contextual & 354 & 3 & 3 & 0 & 0 & 360 \\
Ask-if-uncertain & Updated & 243 & 11 & 3 & 103 & 0 & 360 \\
Ask-if-uncertain & Ambiguous & 128 & 178 & 54 & 0 & 0 & 360 \\
\addlinespace[0.25em]
Full-context & Stable & 358 & 2 & 0 & 0 & 0 & 360 \\
Full-context & Contextual & 354 & 0 & 6 & 0 & 0 & 360 \\
Full-context & Updated & 180 & 0 & 0 & 180 & 0 & 360 \\
Full-context & Ambiguous & 242 & 43 & 66 & 9 & 0 & 360 \\
\bottomrule
\end{tabular}%
}
\end{table}

\begin{table}[H]
\centering
\small
\caption{Policy confusion matrix for Qwen3-32B. Rows are gold benchmark categories; columns are predicted policies plus parse failures.}
\label{tab:app-policy-confusions-2}
\resizebox{\linewidth}{!}{%
\begin{tabular}{llrrrrrr}
\toprule
Agent & Generation category & Use & Ask & Ignore & Update & Parse fail & $n$ \\
\midrule
\textsc{Historical no-state-output} & Stable & 353 & 5 & 1 & 0 & 1 & 360 \\
\textsc{Historical no-state-output} & Contextual & 359 & 1 & 0 & 0 & 0 & 360 \\
\textsc{Historical no-state-output} & Updated & 186 & 16 & 49 & 109 & 0 & 360 \\
\textsc{Historical no-state-output} & Ambiguous & 211 & 119 & 28 & 0 & 2 & 360 \\
\addlinespace[0.25em]
\textsc{Cleaned no-state-output} & Stable & 357 & 3 & 0 & 0 & 0 & 360 \\
\textsc{Cleaned no-state-output} & Contextual & 357 & 3 & 0 & 0 & 0 & 360 \\
\textsc{Cleaned no-state-output} & Updated & 178 & 9 & 52 & 121 & 0 & 360 \\
\textsc{Cleaned no-state-output} & Ambiguous & 231 & 91 & 38 & 0 & 0 & 360 \\
\addlinespace[0.25em]
\textsc{Direct policy (no classification)} & Stable & 319 & 30 & 1 & 10 & 0 & 360 \\
\textsc{Direct policy (no classification)} & Contextual & 350 & 3 & 3 & 4 & 0 & 360 \\
\textsc{Direct policy (no classification)} & Updated & 199 & 35 & 8 & 118 & 0 & 360 \\
\textsc{Direct policy (no classification)} & Ambiguous & 153 & 174 & 9 & 24 & 0 & 360 \\
\addlinespace[0.25em]
\textsc{State-structured bundle} & Stable & 355 & 1 & 1 & 1 & 2 & 360 \\
\textsc{State-structured bundle} & Contextual & 357 & 2 & 1 & 0 & 0 & 360 \\
\textsc{State-structured bundle} & Updated & 41 & 3 & 44 & 271 & 1 & 360 \\
\textsc{State-structured bundle} & Ambiguous & 206 & 104 & 37 & 11 & 2 & 360 \\
\addlinespace[0.25em]
\textsc{Historical oracle-worded routing} & Stable & 359 & 0 & 1 & 0 & 0 & 360 \\
\textsc{Historical oracle-worded routing} & Contextual & 360 & 0 & 0 & 0 & 0 & 360 \\
\textsc{Historical oracle-worded routing} & Updated & 64 & 65 & 51 & 180 & 0 & 360 \\
\textsc{Historical oracle-worded routing} & Ambiguous & 71 & 159 & 127 & 3 & 0 & 360 \\
\addlinespace[0.25em]
Ask-if-uncertain & Stable & 353 & 7 & 0 & 0 & 0 & 360 \\
Ask-if-uncertain & Contextual & 357 & 2 & 1 & 0 & 0 & 360 \\
Ask-if-uncertain & Updated & 283 & 22 & 10 & 45 & 0 & 360 \\
Ask-if-uncertain & Ambiguous & 206 & 133 & 21 & 0 & 0 & 360 \\
\addlinespace[0.25em]
Full-context & Stable & 360 & 0 & 0 & 0 & 0 & 360 \\
Full-context & Contextual & 359 & 0 & 1 & 0 & 0 & 360 \\
Full-context & Updated & 306 & 6 & 11 & 37 & 0 & 360 \\
Full-context & Ambiguous & 275 & 56 & 29 & 0 & 0 & 360 \\
\bottomrule
\end{tabular}%
}
\end{table}

\section{Execution Completeness, Failures, Tokens, and Cost}
The frozen-V3 hosted execution completed 4,800 matched-ablation calls and 5,760 label-conditioning diagnostic calls. One final matched-ablation parse failure remained visible and was counted as incorrect. During an earlier matched-ablation resume attempt, 53 transient provider-failure records were retained in the raw logs; the same call IDs were later completed under the frozen protocol and were not counted as separate valid predictions. Total estimated hosted cost including two smoke calls was \$1.8007 under the authorized \$20.00 hard budget.

Table~\ref{tab:app-call-cost-ledger} separates hosted call accounting by protocol.
\begin{table}[H]
\centering
\small
\begin{tabular}{lrl}
\toprule
Protocol & Calls & Hosted cost \\
\midrule
Matched prompt ablation & 4,800 & \$0.8887 \\
Label-conditioning diagnostic & 5,760 & \$0.9118 \\
Normalized-input probe & 960 & \$0.058766; \$0.066606 \\
Semantic-evidence follow-up & 2,880 & \$0.3977 \\
Smoke tests & 2 & \$0.0002 \\
Frozen-core total including smoke & 10,562 & \$1.8007 \\
\bottomrule
\end{tabular}
\caption{Hosted call and cost ledger by protocol. The normalized-input row lists stored per-model costs rather than a stored protocol-total cost.}
\label{tab:app-call-cost-ledger}
\end{table}

\section{Main-Text Design Materials}
\label{sec:moved-main-materials}
Table~\ref{tab:app-representative-v3-family} reports the representative four-way family.
\begin{table}[H]
\centering
\small
\setlength{\tabcolsep}{4pt}
\begin{tabular}{p{0.08\textwidth}p{0.56\textwidth}p{0.25\textwidth}}
\toprule
Policy & History records & Rule-derived relation \\
\midrule
\textsc{Use} & Jordan saved an earlier entry: flight seat choice uses an aisle seat for its flight seat choice option. The later note records an aisle seat for the flight seat choice option of flight seat choice. Case A: Jordan noted that the train seat option for train seat is a window seat. Case B: the profile lists I sit near an exit as the train seat option for train seat. & One current task-relevant preference is unique; temporal and contextual distractors do not change it. \\
\textsc{Ignore} & The earlier note records a window seat for the train seat option of train seat. Later, Jordan noted that the train seat option for train seat is I sit near an exit. Case A: the profile lists I sit near an exit as the train seat option for train seat. Jordan saved the case b entry: train seat uses a window seat for its train seat option. & The four records concern distractor attributes; no task-relevant memory is needed. \\
\textsc{Update} & Earlier, Jordan noted that the flight seat choice option for flight seat choice is a window seat. Later, the profile lists a bulkhead aisle seat as the flight seat choice option for flight seat choice. Jordan saved the case a entry: train seat uses a window seat for its train seat option. The case b note records I sit near an exit for the train seat option of train seat. & The later temporal task-relevant record supersedes the earlier task-relevant record. \\
\textsc{Ask} & Earlier, the profile lists a window seat as the train seat option for train seat. Jordan saved a later entry: train seat uses I sit near an exit for its train seat option. The case a note records an aisle seat for the flight seat choice option of flight seat choice. Case B: Jordan noted that the flight seat choice option for flight seat choice is a window seat. & Two context-specific task-relevant preferences are present, but the task omits the deciding context. \\
\bottomrule
\end{tabular}
\caption{Representative four-way family from the frozen controlled set. The task is byte-identical for all four variants: ``Choose the preference option for Jordan's flight seat choice.'' The example is shown to expose the controlled synthetic structure, not as a naturalistic conversation sample.}
\label{tab:app-representative-v3-family}
\end{table}

\section{Extended Related Work and Supplementary Discussion}
\label{sec:realm-extended-related}
Long-horizon personalization also raises safety and privacy risks: models can emit training data \citep{carlini2021extracting}, infer private attributes from text \citep{staab2023beyond}, and may require user-level privacy protections \citep{mcmahan2018learning}.

The label-conditioning sensitivity diagnostic uses six artifact conditions: policy-only, state-first, policy-first, forced-correct-state, forced-wrong-state, and deterministic-routing without policy definitions. Forced-state conditions supply a benchmark-implied or deliberately wrong state and ask the model to choose a policy. The wrong state is assigned once before evaluation with seed 20270719 by sampling uniformly from the three non-reference states for each example; the assignment is fixed across models and seeds. These conditions test whether policy routing is sensitive to externally supplied benchmark-associated labels. They do not show that naturally emitted states are faithful internal mechanisms, because the benchmark state labels map deterministically to policy decisions by design.

\textbf{Instability is a diagnostic, not a replacement for confidence calibration.} With three seeds, prediction disagreement is a useful warning signal but not a calibrated probability. We report per-call accuracy for unanimous and non-unanimous examples, majority-vote accuracy for 2--1 splits, and full-disagreement counts. Unanimity is not a guarantee of correctness: many unanimous predictions remain wrong. This supports reporting parse-visible, seed-visible execution traces instead of treating a single sampled output as a stable model behavior.

\textbf{Decomposed evidence fields do not remove the elicitation bottleneck.} The semantic-evidence follow-up avoids supplying the old benchmark-state label, but it does not produce a positive result. Llama performs worse when asked to emit decomposed semantic evidence before the policy decision. GPT-OSS cannot be interpreted cleanly for the explicit-evidence arm because the provider rejected those requests under JSON-mode validation. This strengthens the conservative framing: intermediate-field prompting is an audit target, not a solved mechanism.

\end{document}